\documentclass{article}

\usepackage{microtype}
\usepackage{graphicx}
\usepackage{subcaption}
\usepackage{booktabs} 

\usepackage{hyperref}

\usepackage[preprint]{icml2026}
\usepackage{amsmath}
\usepackage{amssymb}
\usepackage{mathtools}
\usepackage{amsthm}

\usepackage[capitalize,noabbrev]{cleveref}

\theoremstyle{plain}

\theoremstyle{definition}

\theoremstyle{remark}

\usepackage[textsize=tiny]{todonotes}

\begin{document}
\twocolumn[
  \icmltitle{Scalable Machines with Intrinsic Higher Mental-State Dynamics}
  \vspace{0.15cm}

  \begin{icmlauthorlist}
    \icmlauthor{Ahsan Adeel*}{yyy}
    \icmlauthor{M. Bilal}{yyy}
  \end{icmlauthorlist}

  \icmlaffiliation{yyy}{TREND Project: Funded by the UK Advanced Research + Invention Agency (ARIA)} 
  
  \icmlcorrespondingauthor{*Ahsan Adeel}{ahsan.adeel1@stir.ac.uk}
  \begin{center}
\small\textbf{TREND Project:} \texttt{github.com/ARIA-Funded-TREND/IHMS}
\end{center}

  \vskip 0.3in
]

\printAffiliationsAndNotice{}

\textbf{\textit{Drawing on recent breakthroughs in cellular neurobiology and detailed biophysical modeling linking neocortical pyramidal neurons to distinct mental-state regimes, this work introduces a mathematically grounded formulation showing how models (e.g., Transformers) can implement computational principles underlying awake imaginative thought to pre-select relevant information before attention is applied via triadic modulation loops among queries ($Q$), keys ($K$), and values ($V$).~Scalability experiments on ImageNet-1K, benchmarked against a standard Vision Transformer (ViT), demonstrate significantly faster learning with reduced computational demand (fewer heads, layers, and tokens), consistent with our prior findings in reinforcement learning and language modeling. The approach operates at approximately $\mathcal{O}(N)$ complexity with respect to the number of input tokens $N$.}}
\section{Introduction}
Attending to what is relevant is fundamental to both the mammalian brain and machine learning models. Yet determining relevance remains a core challenge. Modern artificial neural networks, such as Transformer models \cite{vaswani2017attention} and their variants, compute relevance through attention over learned $Q$, $K$, and $V$ representations. However, these representations are largely treated as unconstrained learned features and lack intrinsic predictive mechanisms during the feedforward (FF) phase to evaluate whether they are coherent, consistent, contextually appropriate, or even relevant prior to attention.
\\Recent cellular neurobiology \cite{larkum1999new, aru2020cellular, Phillips2024cellular, storm2024integrative} and detailed biophysical modeling \cite{graham2025context} provide a contrasting mechanism. In the mammalian neocortex, layer 5 pyramidal two-point neurons (TPNs) (Figure 1(a)) integrate two distinct input streams: FF sensory evidence (receptive field; RF\footnote{RF refers to the external world: the region of the sensory periphery where stimuli influence the electrical activity of sensory cells.}, denoted $R$) arriving at basal dendrites, and contextual input (contextual field; CF\footnote{CF refers to the internal world: signals from diverse cortical and subcortical sources, including feedback.}, denoted $C$) arriving at apical dendrites. When basal and apical compartments are simultaneously depolarized, TPNs generate brief high-frequency bursts, signaling coherence between evidence and context. This bursting minimizes predictive error, or free energy \cite{friston2010free}, by selectively amplifying relevant information while suppressing irrelevant signals \cite{marvan2024cellular}, enabling context-sensitive processing before attention.
\\Bursting probability depends on the relative strengths of $R$ (evidence) and $C$ (context) inputs and varies systematically across processing regimes associated with different mental states, including slow-wave (SW) sleep, wakefulness, and rapid eye movement (REM) sleep \cite{aru2020cellular, Phillips2024cellular, storm2024integrative}. Four operational modes of neocortical pyramidal TPNs have been identified \cite{Phillips2024cellular, graham2025context}:
\\(1) During typical wakefulness, apical dendrites ($C$), when receiving moderate input alongside high $R$ input, amplify the transmission of FF basal or perisomatic input. This regime is termed \textit{apical amplification (AA)}, in which bursting probability is primarily determined by $R$ but modulated by $C$. Such context-sensitive cooperative amplification supports cognitive abilities while limiting internally generated imagery and thought.
\\(2) When $C$ input is high and $R$ input is low, apical input can independently drive axonal spiking output, a condition associated with REM sleep and internally generated thought (imagination). This regime is termed \textit{apical drive (AD)}, in which bursting probability is largely determined by $C$ alone. 
\\(3) When both $C$ and $R$ inputs are high, bursting probability is maximal and is associated with imaginative cognition during wakefulness (awake thought); this regime is termed \textit{AD + Awake}. 
\\(4) When $C$ input has no effect on the neuron's current output, the state is associated with SW sleep and is termed \textit{apical isolation (AI)}.
See A.1--A.4 for further details.
\\Inspired by these cellular mechanisms, we developed Cooperative Context-sensitive Cognitive Computation (Co$^4$) \cite{adeel2025beyond}, an architecture that extends standard Transformer models. Co$^4$ leverages internally generated mental-state dynamics, formulated as $R$--$C$ interactions analogous to those described above \cite{Phillips2024cellular, graham2025context}, to enable on-the-fly selection of relevant information prior to attention. These dynamics are implemented through triadic modulation loops among $Q$-, $K$-, and $V$-like populations modeled on TPNs and governed by modulatory (MOD) cooperation laws associated with mental-state regimes \cite{graham2025context}.
\\The MOD cooperation laws used here follow the qualitative structure of dendritic transfer functions reported in biophysical models of TPNs \cite{graham2025context}, where the functional form matters more than the exact analytic specification. They mirror nonlinear $R$–$C$ interactions and represent bursting probability as a continuous amplitude reflecting the coherence between the $R$ and $C$ streams. This provides an empirically grounded basis for the proposed regime-dependent dynamics. These regimes, however, serve as operational descriptors and do not imply that Co$^4$ possesses subjective experience or consciousness.
\\During triadic modulation, each TPN population treats its latent representation as the primary evidence stream ($R$), while receiving contextual input ($C$) from the other evolving populations. In Q-TPNs, latent $Q$ is refined through contextual signals from evolving $K$, amplifying those $Q$ elements coherent with $K$ and progressively shaping a context-sensitive set of queries. Similarly, in K-TPNs, latent $K$ is updated in response to evolving $Q$, selectively strengthening those $K$ components aligned with the emerging query structure and refining the contextual representation. In V-TPNs, latent $V$ is stabilized through interaction with the jointly updated $Q$ and $K$, reinforcing values that are consistent with both evidence and contextual structure. Processing regimes for individual Q-, K-, and V-TPNs shift between AA, AD, and AD + Awake depending on the relative strengths of $R$ and $C$ inputs. 
\\Once Co$^4$ establishes coherence, i.e., separates coherent (relevant) from incoherent (irrelevant) Q--K--V token representations, the exact downstream readout operator (e.g., pruning, gating, or an MLP in place of attention) becomes less critical and can be selected based on design priorities. Applying attention only to the top-$k$ relevant tokens operates at $\mathcal{O}(N + k^2)$ complexity, where $k \ll N$ and $k \leq \sqrt{N}$, resulting in near-linear scaling with respect to the number of tokens $N$. Replacing attention with a simple MLP applied to the modulated $V$ 
yields strictly $\mathcal{O}(N)$ complexity.
\\We evaluate Co$^4$ on ImageNet-1K, Mini-ImageNet, Tiny-ImageNet, CIFAR-10, and reinforcement learning (RL) benchmarks. Across tasks, Co$^4$ achieves significantly faster learning with reduced computational demand than a standard Transformer model under the same conditions.

\section{Background}
The flexible integration of top-down contextual signals $(C)$ and bottom-up sensory evidence $(R)$ provides a cellular substrate for several computational theories of neocortical function \cite{aru2020apical, bachmann2020dendritic, marvan2021apical, aru2020cellular}. One prominent framework for such dual-stream processing is predictive coding and related formulations grounded in the principle of free-energy minimization \cite{friston2010free, friston2005theory, friston2018deep, friston2017graphical}. In predictive coding models, predictions $(C)$ generated at higher cortical levels are compared with incoming sensory signals $(R)$, such that inputs consistent with the predictions are attenuated while mismatches propagate upward as prediction errors. However, accumulating evidence \cite{marvan2024cellular} suggests that $R$–$C$ interactions are not limited to purely subtractive comparison. Instead, convergent inputs can interact nonlinearly, such that coherent contextual and sensory signals cooperatively amplify neuronal responses through bursting. Work on burst-dependent synaptic plasticity \cite{payeur2021burst} further indicates that these dendritic $R$–$C$ interactions support coordinated credit assignment across hierarchical circuits, providing a biologically plausible approximation to gradient-based optimization. However, although such approaches are inspired by TPN mechanisms for learning, their processing dynamics do not directly reflect TPN computation. More broadly, flexible $R$–$C$ coupling has been linked to integrative processes associated with conscious processing \cite{aru2020cellular, storm2024integrative, marvan2021apical} and to FF processing across distinct mental-state regimes \cite{Phillips2024cellular, graham2025context}. These findings motivate the MOD dynamics employed in $\mathrm{Co}^4$. See A.3 for extended background.
\section{$\rm{Co}^4$ Architecture}
The Co$^4$ mechanism (Figure 1(b)) dynamically tunes latent $Q$, which acts as evidence ($R$) (i.e., \textit{what should be looked for}), using contextual input ($C$) from the evolving $K$ and $V$. Similarly, latent $K$, acting as $R$, evolves in response to the current $Q$ (i.e., \textit{what is being asked}) and the associated values $V$, which together provide contextual input ($C$). In turn, $V$, acting as $R$, adapts through interaction with the updated $Q$ and $K$, which together form its contextual input ($C$). In effect, each internally generated token implicitly evaluates:
\begin{quote}
\textit{“How should I represent myself given what this latent is attempting to understand?”}
\end{quote}

This reciprocally adaptive loop is implemented through the cooperation of Q-, K-, and V-TPN-inspired populations. In this process, Q-TPNs generate modulated queries $Q_m$ by using latent queries $Q_{\mathrm{L}}$ as $R$, and input projections $Q_X$ and $K_X$, together with the internal belief state $\mu$, as $C$. Here, $Q_X$ serves as proximal context (P), $K_X$ as distal context (D), and $\mu$ as universal context (U), representing a local joint prediction error between evolving latent representations and sensory evidence. The contextual field $C$ is therefore not monolithic but consists of multiple sources that collectively modulate the evidence stream $R$, analogous to the integration of diverse contextual inputs across dendritic compartments in pyramidal neurons \cite{adeel2022unlocking, pagkalos2023leveraging}. This formulation allows $Q_m$ to dynamically align with both sensory evidence and emerging internal hypotheses.
\\Similarly, K-TPNs generate modulated keys $K_m$ by using latent keys $K_{\text{L}}$ as $R$, and input projections $Q_X$ and $K_X$, together with $\mu$, as $C$. In this configuration, $K_X$ serves as P, $Q_X$ as D, and $\mu$ as U. Accordingly, $K_m$ is refined in coordination with evolving internal and sensory signals.
\\V-TPNs generate modulated values $V_m$ by using latent values $V_{\text{L}}$ as $R$, and input projection $V_X$, together with $Q_m$, $K_m$, and $\mu$, as $C$. Here, $V_X$ serves as P, $Q_m$ and $K_m$ as D, and $\mu$ as U. Accordingly, $V_m$ evolves through interaction with the jointly updated $Q_m$ and $K_m$, encoding coherent hypotheses grounded in both internal inference and sensory validation.
\\For each raw latent token ($Q_{\text{L}}$, $K_{\text{L}}$, $V_{\text{L}}$), Co$^4$ generates distinct sets of $(Q_m, K_m, V_m)$, each undergoing customized triadic interaction. The resulting coherent Q--K--V representations are then pooled for top-$k$ token attention, or alternatively, only $V_m$ is passed to an MLP layer.
\\The formal equations below describe the case in which $Q_L$, $K_L$, and $V_L$ are initialized from normal distributions. In this setting, strong contextual inputs from the input projections place the system in the AD regime.\\
The update rule for the internal joint belief state $\mu$, which governs prediction success maximization \cite{marvan2024cellular} in Eqs. (7--9) and can be interpreted as a precision-like contextual gain signal in predictive-coding terms, is given by:
\begin{equation}
\mu \leftarrow \mu \odot (1 + \alpha E),
\end{equation}
where $\alpha$ is the step size, $\odot$ denotes elementwise multiplication, and $E$ represents the aggregated local prediction error that compare latent predictions with modulated representation to update belief state:
\begin{equation}
E = |\epsilon_q| + |\epsilon_k| + |\epsilon_v|,
\end{equation}
with individual prediction errors given by:
\begin{equation}
\epsilon_q = Q_m - Q_{\text{L}}, \quad
\epsilon_k = K_m - K_{\text{L}}, \quad
\epsilon_v = V_m - V_{\text{L}}
\end{equation}
The MOD cooperation laws (Eqs. 4–6), together with the computations of $Q_m$, $K_m$, and $V_m$ defined in different formulations (Eqs. 7–12), are grounded in burst-probability transfer functions (Eqs. 20–23, see A.4) derived for TPNs operating in distinct processing regimes (AA, AD, AD+Awake) \cite{graham2025context}, consistent with the dendritic transfer functions reported in \cite{kay2020contextual, kay2022comparison}. In this formulation, bursting probabilities are translated into a continuous amplitude that reflects the degree of coherence between the $R$ and $C$ streams during the FF phase. The resulting MOD functions therefore define a structured cooperation surface over the $R$–$C$ variables, shaping representational salience and guiding gradient flow along coherent evidence–context trajectories. These functions can be expressed in the following general forms:
\begin{equation}
MOD_1(R, C_1, C_2) = f_c(C_1) + g(R, C_2)
\end{equation}
\begin{equation}
MOD_2(R, C) = f_r(R) + g(R, C)
\end{equation}
\begin{equation}
MOD_3(R_1, R_2, C) = f_r(R_1) + f_c(C) + g(R_2, C) 
\end{equation}

where $C_1$, $C_2$, $R_1$, $R_2$ represent possible combinations of different types of receptive and contextual fields, including proximal, distal, and universal inputs \cite{adeel2020conscious, adeel2022unlocking}, and may vary depending on the formulation used (e.g., see Eqs. 7–9). Generally, $f_c(C)$ represents the unilateral integrated contextual drive function, $f_r(R)$ denotes the unilateral integrated evidence function, and $g(R, C)$ represents cooperative interaction, modeled here as a bivariate multiplicative function (alternative interaction forms are possible). Larger values of $\text{MOD}(R, C)$ correspond to highly salient latent representations.
\\Eq. 4 represents the AD regime and captures high bursting under low $R$ and high $C$ inputs. Since this MOD form lacks an independent evidence-driven term $f(R)$, it encodes a context-governed inductive bias: evidence influences cooperation only through a context-controlled gate, corresponding to high MOD activation, linked to imaginative or internally generated cognition. Consequently, a purely evidence-dominant regime, such as AA is not achievable within this formulation. 
\\Eq. 5 represents the AA regime and captures moderate bursting under high $R$ and moderate $C$ inputs. Salience is primarily driven by sensory evidence $R$, while context $C$ modulates amplification without dominating. This corresponds to moderate activation supporting perception.
\\Eq. 6 represents the AD + Awake regime and captures maximum bursting under high $R$ and high $C$ inputs. When both $R$ and $C$ are large, the system enters a maximal coherence regime. In this maximal regime, both unilateral drives and their cooperative interaction are active simultaneously. Both evidence and context independently drive amplification. In this regime, both sources can independently initiate bursting, and their interaction further enhances amplification. This MOD activation supports contextually grounded imagination and inference. 
\\The internally generated (predicted) latent queries are defined as follows:
\begin{equation}
Q_m = (Q_X + \mu) + Q_L (K_X + \mu)
\end{equation}

This is a MOD function of the form shown in Eq. 4, with
\[
R = Q_L, C_1 = Q_X + \mu, C_2 = K_X + \mu
\]
where $C_1$ is an additive combination of the proximal and universal contexts, and $C_2$ is an additive combination of the distal and universal contexts. 


Similarly, the modulated keys are given by:

\begin{equation}
K_m = (K_X + \mu) + K_L (Q_X + \mu)
\end{equation}

This is a MOD function of the form shown in Eq. 4, with
\[
R = K_L, C_1 = K_X + \mu, C_2 = Q_X + \mu
\]
where $C_1$ is an additive combination of the proximal and universal contexts, and $C_2$ is an additive combination of the distal and universal contexts.

The evolution of latent value tokens is given by:
\begin{equation}
V_m = V_X^2 + 2V_X + (Q_m + \mu)(K_m + \mu)(1 + |V_L|)
\end{equation}

This asynchronous MOD function follows a form similar to that proposed in \cite{adeel2022unlocking}\footnote{The functional form was originally identified empirically in prior audio-visual speech enhancement experiments, where visual context ($C$) modulated and could dominate noisy auditory evidence ($R$) under low signal-to-noise ratios. The resulting interaction structure captures context–evidence cooperation and is consistent with the nonlinear transfer functions reported in biophysical models of pyramidal neurons \cite{graham2025context}.} but here context drives the interaction due to randomly initialized latents; the specific formulation shown in Eq. 4, with
\[
R = 1 + |V_L|, C_1 = V_X^2 + 2V_X, C_2 = (Q_m + \mu)(K_m + \mu)
\]
where $C_1$ is a quadratic function of the proximal context, and $C_2$ is a composite function of the distal contexts $(Q_m, K_m)$ and the universal context $\mu$.
In these MOD functions, when $C$ (i.e., $C_1$ and $C_2$) $\approx 0$, evidence $R$ has no influence on cooperation. When both $R$ and $C$ have minimal strengths, they cooperate weakly, a regime termed Apical Cooperation (AC) \cite{graham2025context}. When $|C|$ is moderate to large and $|R|$ is small, cooperation is dominated by contextual self-drive, yielding context-led salience generation consistent with belief-driven or internally generated interpretations, i.e., the AD regime. Overall, in this MOD form, $C$ determines whether $R$ is amplified or attenuated. Figure 1(c) illustrates a general form of the MOD function for the case of a single $C$ and $R$. 
\\When latents are not initialized from a normal distribution but instead are set via input projections such that $Q_L$ = $Q_X$, $K_L$ = $K_X$, $V_L$ = $V_X$, the resulting $Q_m$, $K_m$, and $V_m$ take the following forms:
 
\begin{equation}
Q_m = Q_L + Q_L K_X
\end{equation}

This is a MOD function of the form shown in Eq. 5, with
\[
R = Q_L, \quad C = K_X
\]


\begin{align}
K_m &= K_L + K_L Q_X
\end{align}

This is a MOD function of the form shown in Eq. 5, with
\[
R = K_L, \quad C = Q_X
\]

In this formulation, $Q_m$ and $K_m$ operate in AA regime in which activation is determined by $R$ but with modulation by $C$. 

\begin{equation}
V_m = \mathrm{ReLU}_\alpha(V_L^2 + 2V_L + Q_m K_m (1 + |V_L|))
\end{equation}
This is a MOD function of the form shown in Eq. 6, with
\[
R_1 = V_L^2 + 2V_L, R_2 = 1 + |V_L|, C = Q_m K_m
\]

The $\mathrm{ReLU}_{\alpha}(x)$ function is a variant of the Rectified Linear Unit (ReLU) activation function commonly used in deep learning models and is defined as:

\[
\mathrm{ReLU}_{\alpha}(x) = \min\!\big(\max(0, x),\, \alpha \big)
\]
\begin{figure*}
  \centering
  \begin{subfigure}[t]{0.3\textwidth}
    \centering
    \includegraphics[width=\textwidth]{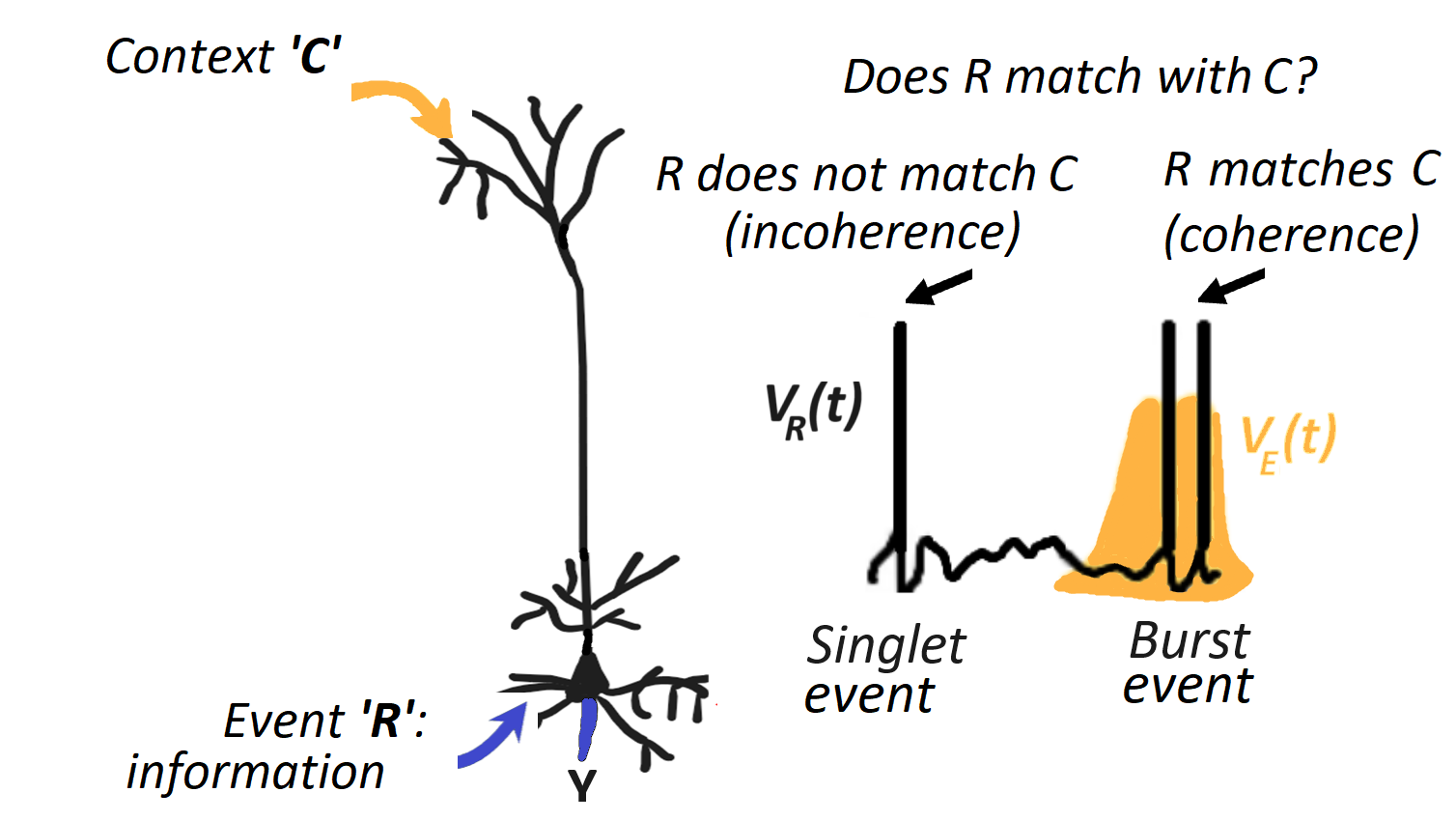}
    \caption*{(a) A pyramidal two-point neuron \cite{larkum1999new, Phillips2024cellular} integrates receptive field $(R)$ input at the basal dendrites and contextual field $(C)$ input at the apical dendrites. The contextual field includes inputs from diverse sources, including feedback. Their co-activation induces burst firing and selective amplification of coherent signals.}
    \label{fig:tpn}
  \end{subfigure}
  \hfill
   \begin{subfigure}[t]{0.68\textwidth}
    \centering
    \includegraphics[width=\textwidth]{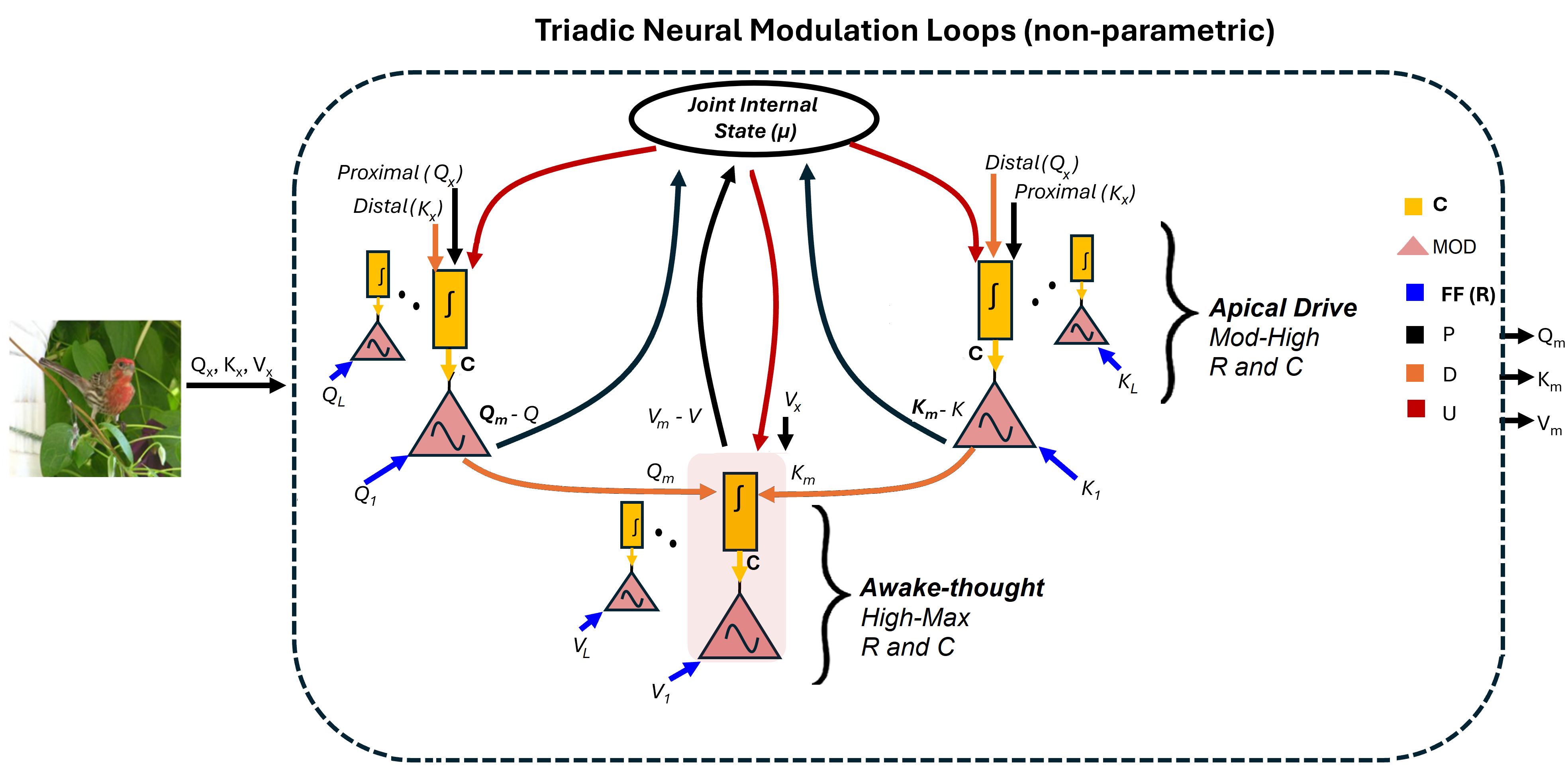}
    \caption*{(b) Co$^4$ architecture: latent $Q_{\text{L}}$, $K_{\text{L}}$, and $V_{\text{L}}$ tokens are initialized from a random distribution and serve as feedforward (FF) inputs i.e., receptive fields (R). Contextual input, including $Q_X$, $K_X$, and $V_X$ are FF contextual projections that assume proximal (P) or distal (D) roles relative to the active TPN population, whereas $\mu$ functions as universal (U) context and provides the recurrent feedback within the iterative dynamics. The TPN-like circuits governing $Q_m$, $K_m$, and $V_m$ evolve via MOD dynamics determined by the relative magnitudes of $R$ and $C$ inputs, corresponding to the neurobiological apical drive (AD) and AD + awake processing regimes. The resulting modulated representations $Q_m$, $K_m$, and $V_m$ are subsequently selected and passed to the self-attention block, or alternatively, $V_m$ is simply passed to an MLP layer. See A.5 for the basic architecture.}
    \label{fig:co4}
  \end{subfigure}
  \vspace{0.7em}
  \begin{subfigure}[t]{0.49\textwidth}
    \centering
    \includegraphics[width=\textwidth]{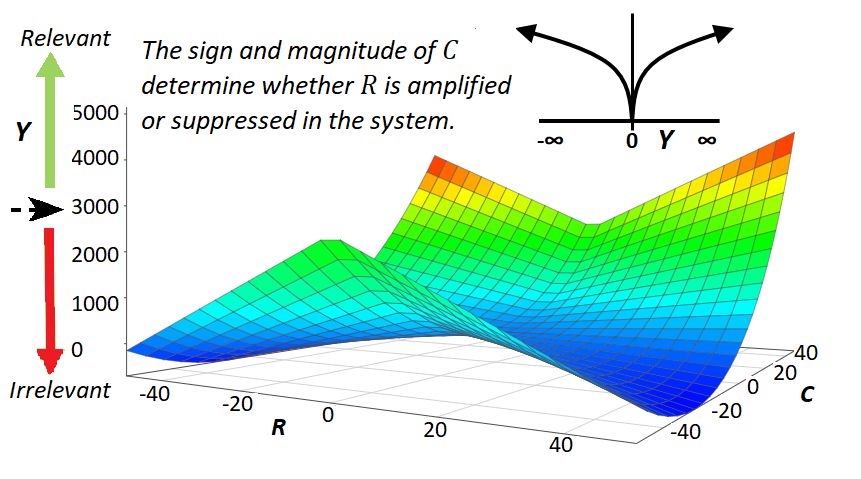}
    \caption*{(c) MOD function: $MOD (R, C) = C^2 + 2C + C(1 + |R|)$}
    \label{fig:mod}
  \end{subfigure}
  \hfill
  \begin{subfigure}[t]{0.49\textwidth}
    \centering
    \includegraphics[width=\textwidth]{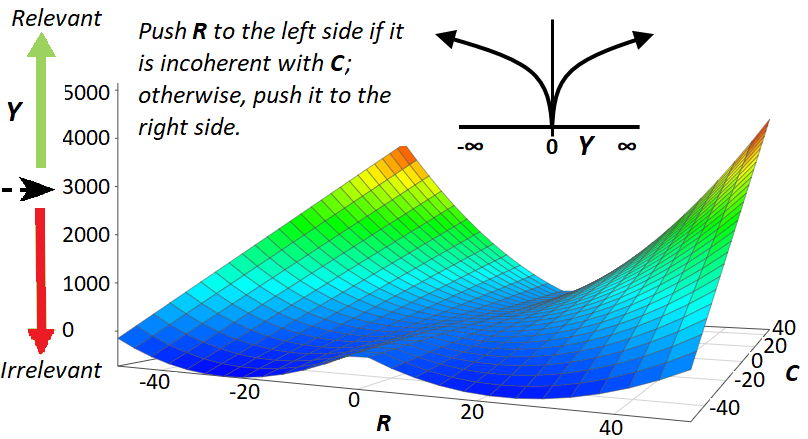}
    \caption*{(d) MOD function: $MOD (R , C) = R^2 + 2R + C(1 + |R|)$}
    \label{fig:contour}
  \end{subfigure}
   \vspace{0.7em}
  \begin{subfigure}[t]{0.45\textwidth}
    \centering
    \includegraphics[width=\textwidth]{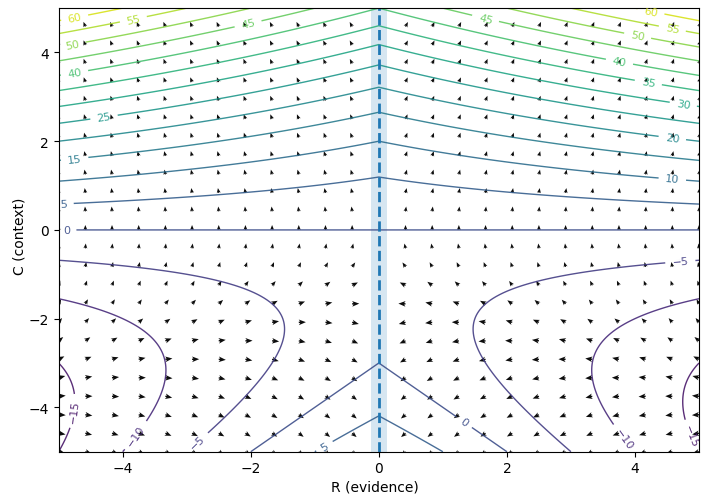}
    \caption*{(e) Vector field visualization and contour plots: $MOD (R, C) = C^2 + 2C + C(1 + |R|)$}. 
    \label{fig:mod}
  \end{subfigure}
  \hfill
  \begin{subfigure}[t]{0.45\textwidth}
    \centering
    \includegraphics[width=\textwidth]{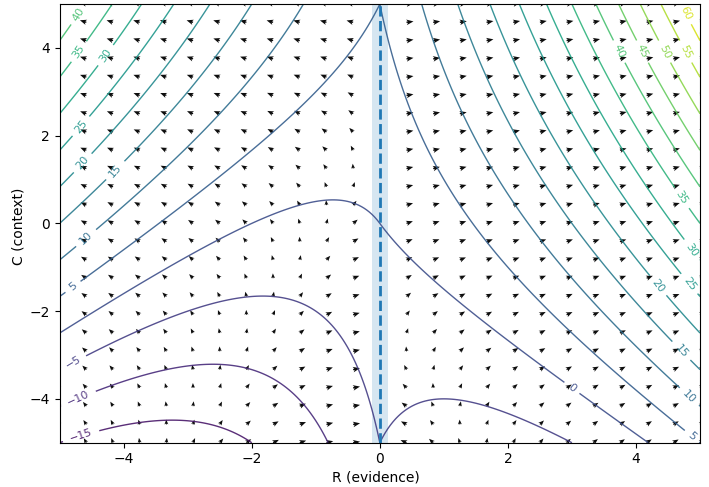}
    \caption*{(f) Vector field visualization and contour plots: $MOD (R, C) = R^2 + 2R + C(1 + |R|)$}.
    \label{fig:contour}
  \end{subfigure}
  \caption{
   Integrated view of biologically inspired mechanisms: (a) Pyramidal two-point neuron; (b) $\rm{Co}^4$ triadic reasoning via $Q$–$K$–$V$ interactions; (c–d) MOD function dynamics illustrating context-sensitive filtering; $Y$ represents the FF signal, which is separated into relevant and irrelevant streams depending on the strength of $C$; (e–f) MOD contours and vector fields across the $R$–$C$ space. Surface plots are presented over a wider range of $R$ and $C$ to illustrate the global geometry of the cooperation landscape, while contour and vector-field visualizations focus on a smaller range to highlight local gradient flow and regime transitions near the origin. Variations in the strengths of $R$ and $C$ shift the system across distinct processing regimes analogous to the neurobiological AA, AD, and AD+Awake regimes, producing corresponding geometric deformations in gradient flow. By shaping representations prior to downstream readout, these modulation laws guide optimization along $R–C$ interaction manifolds, reducing propagation through noisy or irrelevant directions.}
  \label{fig:fig1-combined}
\end{figure*}
\\where $\alpha >0$ is the configurable maximum activation value, $\alpha = 6$ is used in the experiments reported here.
\\In EQ. 12, $C$ as a `modulatory force' pushes the information to the positive side of the activation function (e.g., ReLU6) if $R$ is relevant, otherwise to the negative side. In essence, strong $C$ can discourage or encourage amplification of neural activity regardless of $R$'s strength (strong or weak). The value of $C$ captures the strength of the signal, but different approximate ranges of this value are distinctly associated with different mental-state-dependent processing regimes. The differing behaviour reflects the nonlinearity in the interaction between $C$ and $R$. Eq. 12, visualized in Figure 1(d) as the general form of the MOD function for the case of a single $C$ and $R$, includes: an independent evidence term $f(R) = R^2 + 2R$ and an interaction term $g(R,C) = C(1 + |R|)$. This enables nuanced behavior across regimes: when both $R \approx 0$ and $C \approx 0$, cooperation is minimal. When $R$ dominates and $C$ is weak, the evidence term $f(R)$ controls the salience amplitude and gradients. This reflects an AA-like regime, where perception is driven by strong sensory evidence with weak contextual bias. When both $R$ and $C$ are strong, the system achieves large amplitude and well-directed gradients. This corresponds to the AD + Awake regime of grounded imagination, where context actively tests and refines evidence for maximal coherence. Eqs. (10-11), by contrast, reflect a simpler, additive cooperation model. This supports a perception-like AA regime, where context modulates but does not dominate evidence. See A.7 for spiking simulations of context-sensitive pyramidal neurons implementing interaction dynamics similar to Eq. 12, where bursting probabilities across varying strengths of $R$ and $C$ exhibit regime transitions.
\\The specific form of the triadic modulation loops and the $R$–$C$ integration strategies may vary across datasets and hyperparameter configurations. For example, Figure 1(b) represents one of the architectural variants evaluated in this paper, building on the basic architecture (see A.5); other architectural instantiations are possible. The key element, however, is the cooperative modulation dynamics ($\mathrm{MOD}(R, C)$) operating under distinct mental-state-dependent processing regimes \cite{Phillips2024cellular, graham2025context}. 
\subsection{Gradient Flow}
Rather than assigning global credit exclusively during the feedback (FB) phase via standard backpropagation to reduce prediction error, $\mathrm{Co}^4$ introduces local, credit-like contextual signals from neighboring neurons during the FF phase. These signals, together with joint local prediction error represented by the internal state $\mu$, allow representations to be adaptively refined on the fly prior to global supervision. This constitutes an unsupervised representational shaping phase that precedes task-driven weight updates. Global FB through standard backpropagation is then used to fine-tune connection weights with respect to the task objective. This hybrid local–global mechanism facilitates the rapid emergence of relevant features at early stages, contributing to faster learning.
\\$\rm{Co}^4$ achieves gradient stability by shaping the underlying \emph{R--C} response surface through triadic modulation prior to the downstream readout. The MOD function defines a structured cooperation field over receptive and contextual variables, such that representations are never raw but are generated directly within a context-conditioned dynamical phase space. As a result, gradients propagate along coherent trajectories defined by the geometry of the cooperation field rather than through entangled input space.
\\From a dynamical systems perspective, the vector field (Figure 1(e)) induced by Eq. 9 (the general form of the MOD function) corresponds to the gradient of the cooperation surface, where each vector encodes the local direction and magnitude of maximal semantic amplification. Regions of strong contextual coherence produce smooth, directed flows, enabling stable gradient propagation along cognitively meaningful pathways. Conversely, regions of contextual mismatch naturally induce attenuated or saturating dynamics, preventing abrupt transitions and suppressing unstable gradient spikes. However, gradients remain anchored by context, so no learning signal flows through the RF channel when context is off. Hence, learning signals can vanish globally. This cooperation law does not support a genuine AA regime, since evidence cannot drive cooperation or gradients in the absence of contextual input. In Eq. 12, gradients remain anchored by evidence (Figure 1(f), so learning signals rarely vanish globally; both channels can carry gradient. Gradients grow at most linearly in $|R|$ (no exponential blow-up). 
\\Crucially, in the $\rm{Co}^4$ formulation, vanishing and exploding gradients reflect distinct mental regimes of the system. These regimes correspond operationally to low-coherence interaction patterns. Stable learning emerges in regimes, where receptive and contextual streams are coherently aligned and gradients flow along well-conditioned semantic manifolds. Because reasoning is embedded within a single layer and mediated by continuous cooperation dynamics, $\rm{Co}^4$ reduces reliance on architectural depth for constructing relational structure. This shortens effective gradient paths and ensures that learning is governed primarily by the geometry of $R$--$C$ coherence rather than by the accumulation of transformations across deep stacks. \\Hence, $\rm{Co}^4$ stabilises learning not by engineering gates or optimisers, but by reshaping the response surface prior to downstream readout. A possible extension of the current formulation would incorporate global FB (error) signals directly into processing via online learning rules, such as burst-dependent synaptic plasticity \cite{payeur2021burst, Greedysingle}, thereby further integrating local and global credit signals simultaneously.
\begin{figure*}[!t]
    \centering
\includegraphics[width=0.89\textwidth]{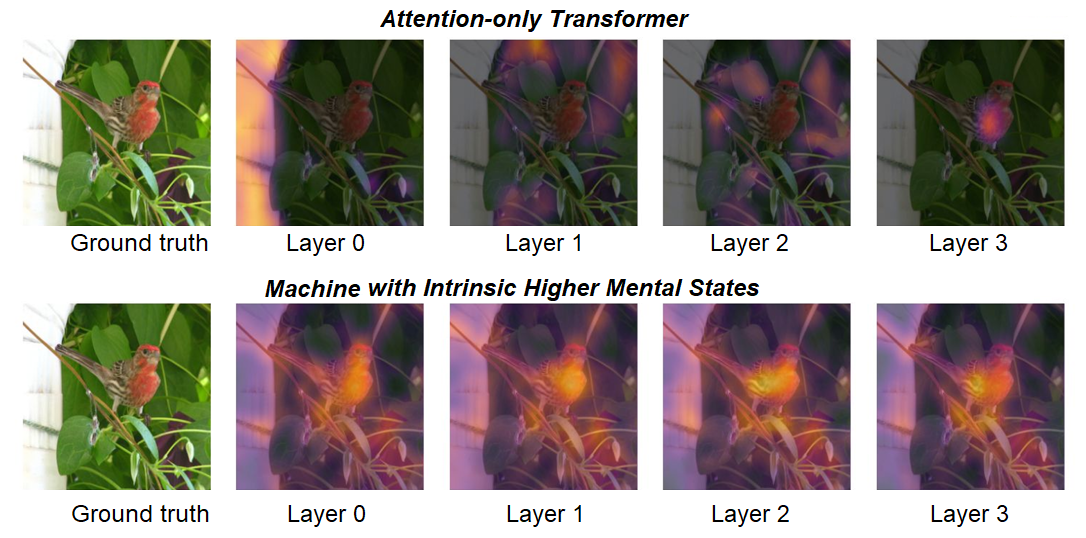}
    \caption{Early training comparison between an attention-only Vision Transformer (ViT) \cite{dosovitskiy2020image}, trained from scratch, and a Co$^4$ machine endowed with intrinsic mental-state-dependent processing regimes analogous to awake imaginative processing \cite{Phillips2024cellular, graham2025context}, which pre-select relevant information \textit{before} attention is applied. The task is to identify a bird from the Mini-ImageNet dataset. Brightness indicates regions emphasized. In the ViT model, this is after attention. In contrast, Co$^4$ rapidly forms a coherent interpretation of the input, highlighting the top-$k$ salient regions via internally generated awake imaginative regimes \textit{before} attention is computed. Co$^4$ exhibits earlier and sharper activation over the semantically relevant object (bird), indicating more coherent internal inference. These findings raise questions about the necessity of deep attention stacks.}
    \label{fig:cifar10}
     \vspace{0.001em}
\end{figure*}
\begin{figure*}[!t]
    \centering
    \includegraphics[width=0.94\textwidth]{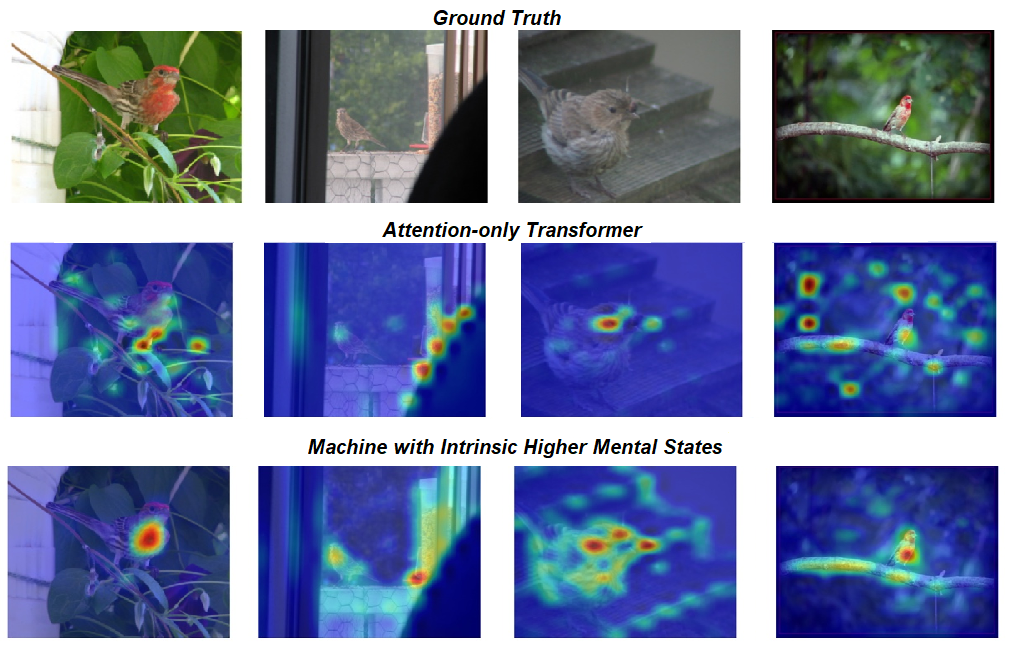}
    \caption{The figure visualizes the complete attention distribution over \textit{N} input tokens: a single-layer Co$^4$ machine versus an attention-only ViT \cite{dosovitskiy2020image}, both trained on Mini-ImageNet for 30 epochs. The ViT exhibits more dispersed attention with less selective localization. In contrast, Co$^4$ demonstrates more centered, context-sensitive activation patterns, indicating stronger spatial coherence.}
    \label{fig:cifar10}
    \vspace{0.001em}
\end{figure*}
\begin{figure*}[t]
  \centering
    \centering
    \includegraphics[width=\textwidth]{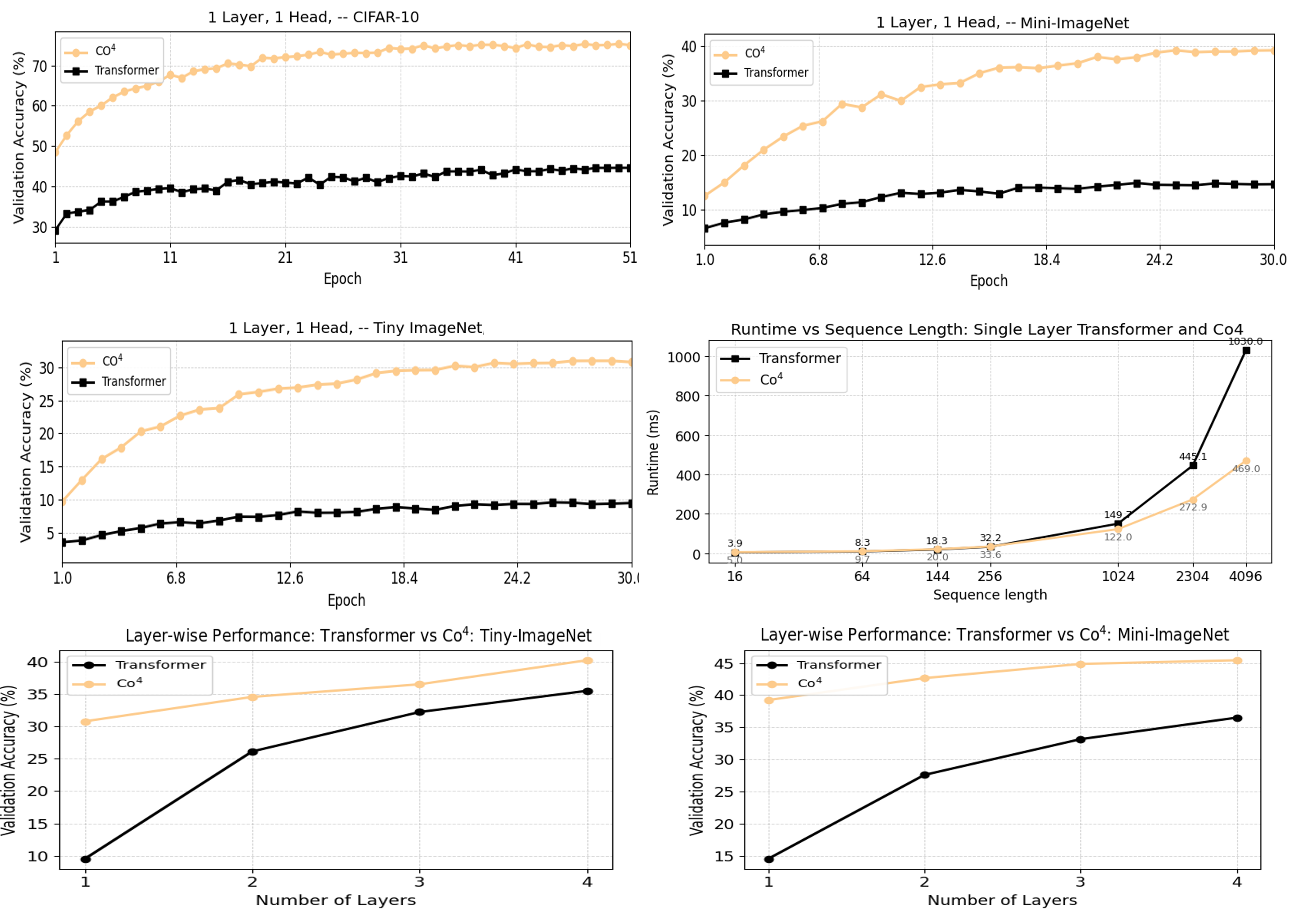}
    \caption{Co$^4$ versus Transformer on CIFAR-10, Tiny-ImageNet, and Mini-ImageNet, trained from scratch: (i–iii) performance of a single-layer model; (iv) inference runtime as a function of sequence length for a single layer (see A.6 for theoretical computational cost analysis); and (v–vi) layer-wise validation accuracy on Tiny-ImageNet and Mini-ImageNet.}
    \label{fig:cifar10}
  \hfill
   \vspace{-1em}
\end{figure*}
\begin{table*}[t]
\centering
\small
\caption{Comparison between Transformer and Co$^4$ models across ImageNet-1K, Mini-ImageNet, Tiny-ImageNet, and CIFAR-10: L, H, Params, and Acc denote the number of layers, attention heads, model parameters, and accuracy, respectively.}
\begin{tabular}{l l c l c c l c c}
\toprule
\textbf{Dataset} & \textbf{Model} & \textbf{L} & \textbf{Combination} & \textbf{Hidden} & \textbf{MLP} & \textbf{H} & \textbf{Params.} & \textbf{Acc.} \\
\midrule
ImageNet-1K & Co$^4$ & \textbf{8} & Co$^4$ & 768 & 3076 & 1 & 53.4M & 70.24\% \\
            & Co$^4$ & 6 & Co$^4$ & 768 & 3076 & 1 & 40.4M & 67.30\% \\
            & \textbf{Co$^4$} & \textbf{6} & \textbf{4} \textbf{Co$^4$ + 2 MHSA} & \textbf{768} & \textbf{3076} & \textbf{1} & \textbf{41.6M} & \textbf{75.01}\% \\
            & Transformer & 12 & MHSA & 768 & 3076 & 12 & 86M & 74.80\% \\
\midrule
Mini-ImageNet & Co$^4$ & 1 & Co$^4$ & 384 & 3076 & 1 & 2.0M & 39.20\% \\
              & Co$^4$ & 4 & Co$^4$ & 384 & 3076 & 1 & 8M & 45.40\% \\
              & Co$^4$-MLP & 1 & Co$^4$ & 384 & 3076 & 1 & 1.87M & 40.21\% \\
              & \textbf{Co$^4$-MLP} & \textbf{4} & \textbf{Co$^4$} & \textbf{384} & \textbf{3076} & \textbf{1} & \textbf{6.07M} & \textbf{48.92}\% \\
              & Transformer & 1 & MHSA & 384 & 3076 & 1 & 2.2M & 14.60\% \\
              & Transformer & 4 & MHSA & 384 & 3076 & 1 & 8.8M & 36.80\% \\
\midrule
Tiny-ImageNet & Co$^4$ & 1 & Co$^4$ & 384 & 3076 & 1 & 1.8M & 30.90\% \\
              & Co$^4$ & 4 & Co$^4$ & 384 & 3076 & 1 & 7.2M & 40.20\% \\
              & Transformer & 1 & MHSA & 384 & 3076 & 1 & 1.9M & 9.50\% \\
              & Transformer & 4 & MHSA & 384 & 3076 & 1 & 7.6M & 35.60\% \\
\midrule
CIFAR-10 & Co$^4$ & 1 & Co$^4$ & 384 & 3076 & 1 & 1.6M & 75.20\% \\
         & Co$^4$ & 4 & Co$^4$ & 384 & 3076 & 1 & 6.4M & 80.30\% \\
         & Transformer & 1 & MHSA & 384 & 3076 & 1 & 1.8M & 44.60\% \\
         & Transformer & 4 & MHSA & 384 & 3076 & 1 & 7.2M & 80.20\% \\
\bottomrule
\end{tabular}
\end{table*}
\begin{table}[t]
\centering
\caption{Effect of varying the number of attention heads on Co$^4$ performance: L, H, Ep, and Acc denote the number of layers, attention heads, epochs, and accuracy, respectively.}
\label{tab:heads}
\begin{tabular}{lccccc}
\toprule
\textbf{Dataset} & \textbf{L} & \textbf{H} & \textbf{Ep.} & \textbf{Embedding} & \textbf{Acc.} \\
\midrule
Imagenet-1k & 6 & 1  & 300 & 768 & 67.30\% \\
            & 6 & 6  & 300 & 768 & 67.81\% \\
            & 6 & 12 & 300 & 768 & 67.89\% \\
Cifar10     & 4 & 1  & 30  & 384 & 80.30\% \\
            & 4 & 4  & 30  & 384 & 80.36\% \\
\bottomrule
\end{tabular}
\end{table}
\begin{table}[t]
\caption{Effect of different modulation equations on CIFAR-10 performance (4 layers, 1 head, 30 epochs, 384-dimensional embeddings).}
\label{tab:mod_eq}
\renewcommand{\arraystretch}{1.3}
\begin{tabular}{lc}
\toprule
\textbf{Modulation Equations} & \textbf{Acc.} \\
\midrule

$\begin{aligned}
Q_m &= Q_X + Q_{\text{L}} \cdot K_X \\
K_m &= K_X + K_{\text{L}} \cdot Q_X \\
V_m &= V_X^2 + 2V_X + Q_m K_m (1 + |V_L|)
\end{aligned}$ 
& 78\% \\
\midrule
$\begin{aligned}
Q_m &= Q_{\text{L}} + Q_{\text{L}} \cdot Q_X \\
K_m &= K_{\text{L}} + K_{\text{L}} \cdot K_X \\
V_m &= V_X^2 + 2V_X + Q_m K_m (1 + |V_L|)
\end{aligned}$ 
&  69\% \\
\midrule
$\begin{aligned}
Q_m &= Q_{\text{L}} + Q_{\text{L}} \cdot K_X \\
K_m &= K_{\text{L}} + K_{\text{L}} \cdot Q_X \\
V_m &= V_X^2 + 2V_X + Q_m K_m (1 + |V_L|)
\end{aligned}$ 
&  72\% \\
\midrule
$\begin{aligned}
Q_m &= Q_X + Q_{\text{L}} \cdot K_X \\
K_m &= K_X + K_{\text{L}} \cdot Q_X \\
V_m &= V
\end{aligned}$ 
& 68\% \\
\midrule
$\begin{aligned}
Q_m &= Q_X + Q_{\text{L}} \cdot K_X \\
K_m &= K_X + K_{\text{L}} \cdot Q_X \\
V_m &= V_X + V_L \cdot Q_m \cdot K_m
\end{aligned}$ 
& 68\% \\
\bottomrule
\end{tabular}
\end{table}
\begin{table}[t]
\centering
\caption{Top-$k$ token selection study comparing Co$^4$ with Transformer (ViT) under identical architectural settings. Performance is shown for multiple values of $k$, together with the ViT baseline \cite{dosovitskiy2020image}: L, Tok, \textit{k}, Acc, and BL denote the number of layers, total tokens, the top-\textit{k} threshold, top-\textit{k} accuracy, and baseline ViT accuracy, respectively.}
\label{tab:topk_compact}
\setlength{\tabcolsep}{4pt}
\renewcommand{\arraystretch}{1.05}
\begin{tabular}{lcccccc}
\toprule
\textbf{Dataset} & \textbf{Model} & \textbf{L} & \textbf{Tok.} & \textbf{$k$} & \textbf{Acc.} & \textbf{BL} \\
\midrule
Tiny-ImageNet 
& Co$^4$ & 4 & 64 & 44 & 31\% & -- \\
& Co$^4$ & 4 & 64 & 32 & 33\% & -- \\
& Co$^4$ & 4 & 64 & 8  & 40\% & -- \\
& Co$^4$ & 4 & 64 & 3  & 37\% & -- \\
& ViT    & 4 & 64 & 8  & 32\% & 35\% \\
& Co$^4$ & 4 & 64 & 8  & 40\% & -- \\
& ViT    & 1 & 64 & 8  & 6\%  & 9\% \\
& Co$^4$ & 1 & 64 & 8  & 35\% & -- \\
\midrule
Mini-ImageNet
& Co$^4$ & 4 & 256 & 128 & 40\% & -- \\
& Co$^4$ & 4 & 256 & 64  & 42\% & -- \\
& Co$^4$ & 4 & 256 & 32  & 43\% & -- \\
& Co$^4$ & 4 & 256 & 16  & 45\% & -- \\
& ViT    & 4 & 256 & 16  & 33\% & 37\% \\
& ViT    & 1 & 256 & 16  & 12\% & 14\% \\
& Co$^4$ & 4 & 256 & 16  & 45\% & -- \\
& Co$^4$    & 1 & 256 & 16  & 16\% & -- \\
\midrule
CIFAR-10 
& ViT    & 4 & 64 & 6 & 76\% & 80\% \\
& Co$^4$ & 4 & 64 & 6 & 80\% & -- \\
\bottomrule
\end{tabular}
\end{table}
\section{Results}
ViT\footnote{\url{https://github.com/google-research/vision_transformer?tab=readme-ov-file}} \cite{dosovitskiy2020image} is used as the baseline for comparison against Co$^4$. Figures 2 and 3 visually demonstrate how Co$^4$ selects relevant information before paying attention. Detailed analysis proceeds as follows: For classification tasks on ImageNet-1K, Mini-ImageNet, Tiny-ImageNet, and CIFAR-10, Co$^4$ with latent initialization from a normal distribution (Eqs. 7–9) is directly compared against the standard Transformer baseline \cite{dosovitskiy2020image}. For classification on Mini-ImageNet, Co$^4$ with an MLP (no attention) and latent initialization from input projection (Eqs. 10–12) is evaluated. For reinforcement learning (RL) tasks, Co$^4$ with latent initialization from input projection (Eqs. 10–12) is benchmarked against the architecture proposed in \cite{tang2021sensory}.
\noindent
\\Across experiments, Co$^4$ consistently achieves superior validation accuracy with significantly fewer layers and attention heads, and converges more rapidly than the standard Transformer using either top-$k$ or simple MLP downstream (no attention) readouts, with computational complexities of $\mathcal{O}(N + k^2)$ and strictly $\mathcal{O}(N)$, respectively. In contrast, the baseline Transformer typically requires deeper stacks, more attention heads, and longer training schedules to reach comparable performance. This performance gap arises because Co$^4$ embeds coherence constraints directly into the representation formation process, allowing the model to identify relevant structures before any downstream readout is applied. Standard attention mechanisms, by contrast, approximate these dynamics only after substantial architectural depth and contextual accumulation.
\\We also observed that single-step Co$^4$ modulation (without $\mu$) achieves performance comparable to its iterative version (with internal belief state $\mu$). In tested environments, a single inference pass appears sufficient to reach a stable representational fixed point. However, in some cases, the iterative Co$^4$ loop, with predictive refinement through an internal state $U$, converged slightly faster to ceiling accuracy. This suggests that coherence can often be established in one step, with marginal gains from further iterations on classification tasks. Additional Co$^4$ iterations may be more beneficial in settings with increased noise, partial observability, or temporally evolving inputs. All results reported below correspond to the single-step modulation setting.
\subsection{Image Classification: ImageNet-1K, Mini-ImageNet, Tiny-ImageNet, CIFAR-10}
Figure 4 and Table 1 present a comparative analysis between ViT and Co$^4$ under same training setup. Despite operating at a lower computational complexity of $\mathcal{O}(12^2)$, where $k=12$ remained constant across all experiments, the Co$^4$ model significantly outperforms the ViT, which incurs a higher complexity of $\mathcal{O}(196^2)$, in terms of learning speed and convergence on the Mini-ImageNet, Tiny-ImageNet, and CIFAR-10 datasets for $N=196$. Moreover, Co$^4$ with a simple MLP downstream module (without attention) performs better than Co$^4$ with top-$k$ attention (Table 1).
\\On large-scale ImageNet-1K, only the top-$k$ variant is evaluated and reported here. An eight-layer $\rm{Co}^4$ model achieves 70.24\% top-1 accuracy with 53.4M parameters. Notably, a hybrid configuration with four $\rm{Co}^4$  layers interleaved with two standard multi-headed self attention (MHSA) layers achieves 75.01\%, compared to a twelve-layer Transformer (74.80\%) while using less than half the parameters (41.6M vs.\ 86M). Testing of Co$^4$ with a simple MLP downstream module (without attention) on ImageNet-1K, and comparisons with newer variants such as Swin-T \cite{liu2021swin}, DeiT-S \cite{touvron2021training}, and VMamba-T \cite{liu2024vmamba}, remain part of ongoing work.

\textbf{Inference runtime} (Figure 4, row 2, column 2)
While both models exhibit comparable inference times at short sequence lengths, where the quadratic cost of attention has not yet become dominant, their runtimes diverge as sequence length increases. Specifically, the Transformer’s inference time grows rapidly, exhibiting the expected quadratic scaling. In contrast, Co$^4$ demonstrates significantly slower growth, approximating linear scaling. This empirically supports the core claim of Co$^4$: by circumventing the quadratic bottleneck of attention, it enables substantially greater scalability for long-context reasoning. A detailed derivation is provided in A.6.

\textbf{Layer-wise comparison} (Figure 4, last row): Compares the layer-wise validation accuracy of a standard attention-only Transformer and the Co$^4$ model with top-$k$ on Tiny-ImageNet and Mini-ImageNet. Across both datasets, Co$^4$ consistently outperforms the Transformer at all depths. Notably, Co$^4$ achieves strong performance even with a single layer, whereas the Transformer exhibits very poor accuracy in the shallow regime and requires additional depth to approach comparable performance. These results indicate that Co$^4$ establishes coherent internal representations with significantly fewer layers than standard Transformers, supporting the claim that intrinsic mental-state-dependent processing regimes enable depth-efficient learning by pre-selecting and amplifying relevant information before attention is applied.

\textbf{Effect of number of heads} (Table 2): Varying the number of attention heads in $\rm{Co}^4$ results in only marginal performance differences across datasets. On ImageNet-1k, increasing the number of heads from 1 to 12 yields an absolute improvement of less than 1\%. This indicates that multi-head attention is not a critical factor for performance in the $\rm{Co}^4$ architecture. This behavior can be attributed to the token modulation and aggregation/generation mechanisms, which enforce strong feature interactions and selective information routing prior to attention. It implies that token modulation may already perform a role analogous to multi-head mixing. This suggests that the proposed architecture is robust to the choice of head count in the tested regimes, allowing a single head to achieve comparable performance with reduced computational overhead. 

\textbf{Regime-collapse ablation studies} (Table 3): This regime-collapse ablation studies show that cooperation laws inspired by mental-state-dependent processing regimes are not arbitrary. Modifying their functional form or nonlinear structure degrades performance, suggesting that preserving the qualitative structure of the MOD formulation is important for achieving optimal results. This provides empirical support for the biological mapping.

\textbf{Top-$k$ token selection and bottleneck effects} (Table 4): Reducing the number of generated tokens via top-$k$ feature leads to improved performance in Co$^4$ whereas the same top-$k$ feature approach in ViT reduces the performance. This sharply contrasts with the typical behavior of Transformer-based models, which often benefit from increased context length. In Co$^4$, however, the opposite trend emerges: \emph{less information leads to better performance}. This provides direct empirical evidence that selective perception can outperform exhaustive perception, supporting the principle that effective reasoning requires internal filtering and compression rather than maximal context exposure.
\begin{figure*} [t]
	\centering
        \includegraphics[trim=0cm 0cm 0cm 0cm, clip=true, width=1\textwidth]{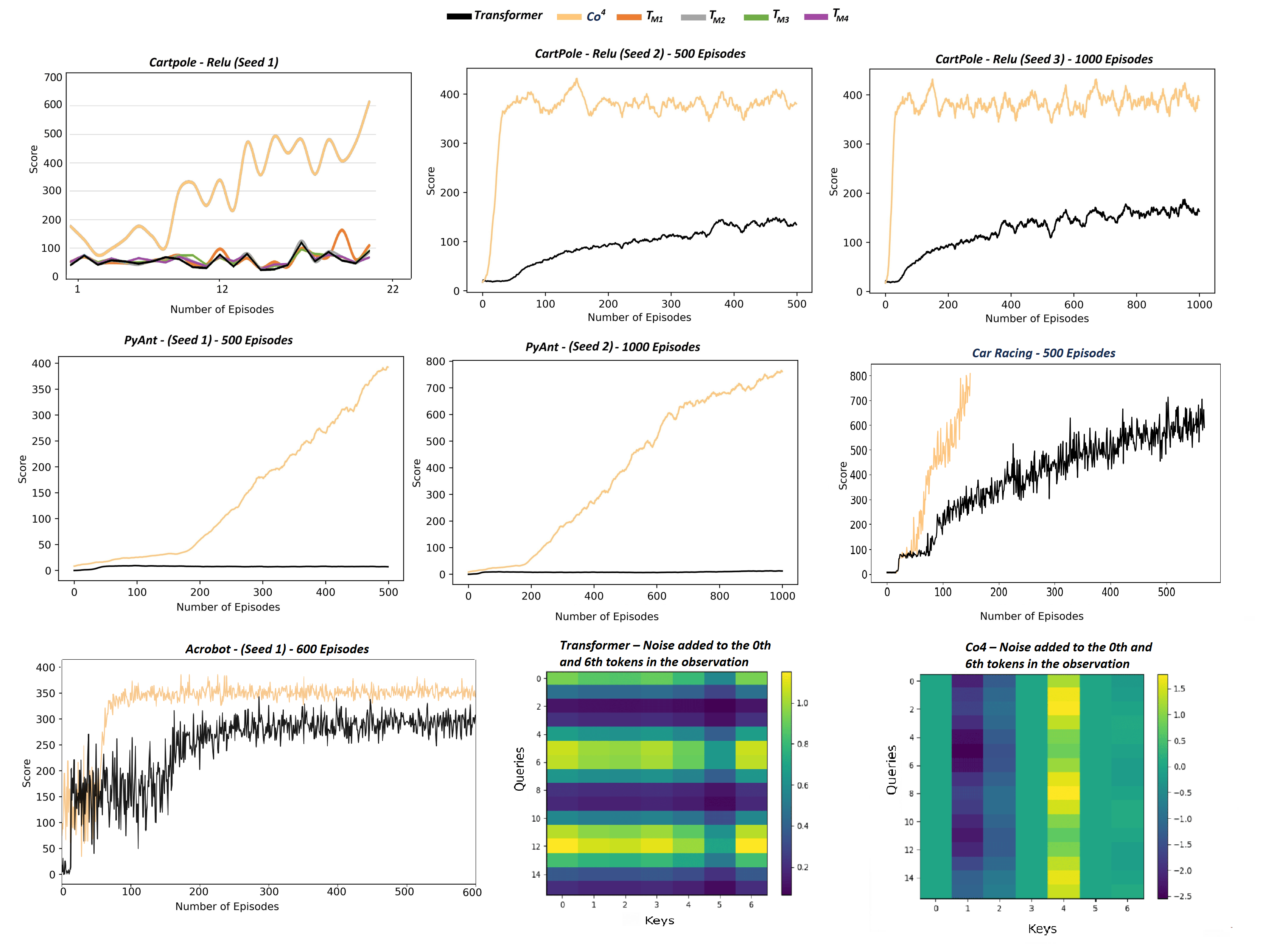}
	\caption{Training results across CartPole (i–iii), PyBullet Ant (iv–v), CarRacing (visual input: $96 \times 96 \times 4$) (vi), and Acrobot (vii), with heat maps (viii–ix) provided as empirical evidence. $T_{M1}$–$T_{M4}$ denote alternative well-established TPN-inspired asynchronous MOD functions \cite{kay2020contextual} (see A.3).
    }
	\label{l5pc}
 \vspace{-1em}
\end{figure*}
\begin{table*}[t]
  \centering
  \begin{minipage}[t]{0.48\textwidth}
    \centering
    \caption{CartPole test scores for 1K to 10K episodes.}
    \label{tab:cartpole}
    \vspace{0.5em}
    \begin{tabular}{lccc}
      \toprule
      \textbf{Model} & \textbf{1K} & \textbf{5K} & \textbf{10K} \\
      \midrule
      Transformer & 279 & 340 & 340 \\
      Trans. (Shuffled) & 279 & 339 & 340 \\
      Co$^4$ & 428 & 524 & 538 \\
      Co$^4$ (Shuffled) & 267 & 508 & 536 \\
      \bottomrule
    \end{tabular}
  \end{minipage}
  \hfill
  \begin{minipage}[t]{0.48\textwidth}
    \centering
    \caption{PyBullet Ant test scores after 1K episodes.}
    \label{tab:pyant}
    \vspace{0.5em}
    \begin{tabular}{lcc}
      \toprule
      \textbf{Model} & \textbf{ES} & \textbf{ES (Shuffled)} \\
      \midrule
      Transformer & 121 & 30 \\
      Co$^4$ & 1170 & 280 \\
      \bottomrule
    \end{tabular}
  \end{minipage}

  \vspace{1.5em}

  \begin{minipage}[t]{0.48\textwidth}
    \centering
    \caption{Acrobot test scores after 200 and 400 training episodes.}
    \label{tab:acrobot}
    \vspace{0.5em}
    \begin{tabular}{lcc}
      \toprule
      \textbf{Model} & \textbf{200} & \textbf{400} \\
      \midrule
      Transformer & 243 & 274 \\
      Co$^4$ & 344 & 350 \\
      \bottomrule
    \end{tabular}
  \end{minipage}
  \hfill
  \begin{minipage}[t]{0.48\textwidth}
    \centering
    \caption{MountainCarContinuous test scores after 50–250 training episodes.}
    \label{tab:mountaincar}
    \vspace{0.5em}
    \begin{tabular}{lccccc}
      \toprule
      \textbf{Model} & \textbf{50} & \textbf{100} & \textbf{150} & \textbf{200} & \textbf{250} \\
      \midrule
      Transformer & 0.58 & 0.21 & 0.41 & 0.21 & 0.23 \\
      Co$^4$ & 0.69 & 0.42 & 0.45 & 0.48 & 0.46 \\
      \bottomrule
    \end{tabular}
  \end{minipage}
\end{table*}
\subsection{Reinforcement Learning}
In the multisensory RL, $V$ and $K$ are functions of the Sensors 1-N i.e., input $x$ $\epsilon$ $R^N$ (e.g., any linear or non-linear transformations) and $Q$ is the output of latent positional encoding. For permutation invariance (PI), the latent $Q$ is independent of the input $x$ so that permutation $x$ only affects $K$ and $V$ but not $Q$, which allows the output to be PI \cite{tang2021sensory}. As explained comprehensively in \cite{tang2021sensory}, the individual sensory inputs 1-N or observations $O_t^i$, \textit{i=1, 2, ... N} along with the previous action $a_{t-1}$ passes through a neural net (NN) module in an arbitrary order such that each NN has partial access to agent's observation at time \textit{t} and $i^{th}$ neuron can only see the $i^{th}$ component of the observation $O_t[i]$, computing $f_K(O_t[i]$, $a_{t-1})$ and $f_V(O_t[i])$ explained in \cite{tang2021sensory}. The overall operation can be described using Eqs. (13-15):
\begin{equation}
K(O_t,a_{t-1})
 =
  \begin{bmatrix}
   f_K(O_t[1], a_{t-1}) \\
   ...\\
 f_K(O_t[N], a_{t-1}) 
   \end{bmatrix} \in  \mathbb{R}^{N \times d_{f_K}} 
\end{equation}
\begin{equation}
V(O_t)
 =
  \begin{bmatrix}
   f_V(O_t[1]) \\
   ...\\
 f_V(O_t[N]) 
   \end{bmatrix} \in  \mathbb{R}^{N \times d_{f_V}} 
\end{equation}
\begin{equation}
\begin{aligned}
m_t = \mathrm{ReLU6}(&R(O_t, a_{t-1})^2 + 2R(O_t, a_{t-1}) \\
&+ C\bigl(1 + |R(O_t, a_{t-1})|\bigr))
\end{aligned}
\end{equation}
\\The architectures of the policy networks, training methods, attention neuron layers, and hyperparameters in all agents are the same as used in \cite{tang2021sensory}. The results presented in Figure 5 were generated using the code provided in \cite{tang2021sensory}, which also serves as the baseline.
\\It was observed that the Co$^4$-driven agent learned tasks significantly faster than the state-of-the-art Transformer-based RL agents \cite{tang2021sensory}. Additionally, the previously proposed context-sensitive neuromodulation transfer functions (Eqs. 16--19 in A.3) performed comparably to the Transformer baseline on the CartPole RL task.
\\Across different random seeds in the CartPole task, Co$^4$ consistently converged to the maximum fitness score in fewer episodes compared to the Transformer baseline. Similar trends were observed in the PyBullet Ant task, the CarRacing task, which uses a high-dimensional visual input space of $96 \times 96 \times 4$, and the Acrobot task \cite{jarnil2024exploring}.
\\Testing results for CartPole (Table 5) show that Co$^4$, trained over 1K–10K episodes, achieved significantly higher fitness scores with much lower standard deviation in both shuffled and unshuffled settings. While Co$^4$ and the baseline performed comparably in the shuffled 1K-episode setting, Co$^4$ quickly surpassed the baseline by 5K episodes, maintaining high performance with reduced variance. 
\\In the PyBullet Ant task (Table 6), Co$^4$ consistently outperformed the baseline across both shuffled and unshuffled scenarios. Similar patterns were also observed in the Acrobot-v1 (Table 7) and MountainCarContinuous \cite{jarnil2024exploring} (Table 8) tasks.
Finally, to provide explainability and empirical insight, attention heatmaps for CartPole visualize the attention scores for both the Transformer and Co$^4$ models under noisy input conditions:
\[
[\textit{Noise},\ x,\ \dot{x},\ \cos(\theta),\ \sin(\theta),\ \dot{\theta},\ \textit{Noise}].
\]
Here, keys and values are generated from the observations, while queries are generated independently. The objective is to identify which observation variables are most relevant for maximizing cumulative reward. After 100 training episodes: The Transformer assigns substantial attention to the injected noise dimensions. The Co$^4$ model effectively suppresses noise, assigning near-zero attention to irrelevant inputs and focusing on the physically meaningful variables. 
\subsection{Empirical Comparison of Regime-Dependent MOD Formulations}

Empirically, different MOD formulations were most effective for different tasks. In CarRacing (RL), Eqs. (10–11) yielded the best performance. In CartPole, PyAnt, Acrobot, and MountainCar, Eq. 12 outperformed the alternatives. For image classification tasks, when latents were initialized from normal distributions, the nonlinear MOD functions (Eqs. 7–9) performed best. When latents were initialized via input projections in Co$^4$ with an MLP (no attention), Eqs. (10–12) yielded superior performance. 
\\These findings suggest that different tasks benefit from different regime-dependent processing dynamics. For example, AA-like regimes appear more suitable for RL tasks where sensory evidence plays a dominant role, whereas AD + Awake–like regimes are more effective when both contextual inference and evidence integration are required. Similarly, in image classification, when latents are initialized from a normal distribution, Eqs. (7–9) corresponding to the AD regime are more suitable, as the latents do not directly represent the input ($X$), which is instead incorporated through strong contextual signals.
\\In contrast, when latents are initialized directly from input projections ($X$), Eqs. (10–12), corresponding to AA and AD + Awake regimes, are more appropriate since both evidence and context provide strong signals. These empirical results indicate that no single MOD formulation is universally optimal. Rather, different regime-dependent dynamics are better suited to different task domains, suggesting that regime-dependent processing provides task-adaptive inductive biases rather than enforcing a fixed operating mode.

\textbf{Code Availability:} The implementation of the multi-layered Co$^4$ model and all results are reproducible using the released open-source code, which includes training scripts with hyperparameters, evaluation pipelines, configuration files, and pretrained checkpoints.

\section{Limitations and Conclusion}

Experiments conducted under constrained computational resources demonstrate that Co$^4$ can substantially accelerate learning while reducing computational demands. By embedding context–evidence cooperation directly into representation formation, the architecture enables models to identify relevant information prior to downstream readout, reducing reliance on architectural depth and quadratic attention scaling. While scaling to ImageNet-1K indicates promising potential, the full compound effects of the Co$^4$ architecture remain to be systematically evaluated. Preliminary experiments using Co$^4$ with a simple MLP as a downstream readout, operating at $\mathcal{O}(N)$ computational cost, indicate the potential for further improvements in both effectiveness and efficiency. Ongoing evaluations on larger datasets aim to clarify these trends and will be reported in future work.
\\Realizing the full design space of Co$^4$ requires rigorous investigation of alternative contextual integration strategies, diverse triadic modulation loop configurations, and joint belief update mechanisms, informed by principles from both modern machine learning and cellular neurobiology. Such exploration is computationally demanding; therefore, we release the framework as an open research platform to facilitate community-driven investigation and iterative refinement of the core mechanism. Whether Co$^4$ improves asymptotic scaling behavior at very large compute budgets remains an open question.
\\The initial evidence presented here suggests that architectures inspired by the cellular foundations of higher mental states may provide a promising path toward deep, deliberate imaginative reasoning in artificial systems. This approach opens the door not only to implementing large numbers of lightweight, inference-efficient AI modules but also to moving these systems beyond mere information processing toward contextual reasoning, shifting from raw efficiency to real understanding. Overall, emulating context--evidence cooperation observed in cortical computation could represent a conceptual step toward more cognitively meaningful machine intelligence.


\section*{Impact Statement}
Whether such advancements are used for beneficial or harmful purposes will ultimately depend on human choices. We hope that the prosocial capacities that have shaped human societies \cite{bregman2020humankind} will guide their development toward the benefit of society. For broader impact, see A.8.

\section{Acknowledgments}
Advanced Research + Invention Agency (ARIA): Nature Computes Better Opportunity seeds. Cooperation is All You Need \cite{adeel2023cooperation} Team. TREND project team (https://cmilab.org/aichip/team/), PhD students and postdocs, including Mohsin Raza, Talha Bin Riaz, Noor Ul Zain, Eamin Ch, Reyhane Ahmadi, Khubaib Ahmed. Professor Bill Phillips, Professor Leslie Smith, and Professor Bruce Graham from the University of Stirling. Dr James Kay from the School of Mathematics, University of Glasgow. Professor Johan Frederik Storm (Professor in Neurophysiology) from the University of Oslo. Professor Panayiota Poirazi from IMBB-FORTH. Professor Newton Howard from Oxford Computational Neuroscience. Professor John Broome (Emeritus White’s Professor of Moral Philosophy) from Corpus Christi College, University of Oxford. Professor Peter König from the University Osnabrück. Professor Heiko Neumann from Ulm University, and several other eminent scholars for their help and support in several different ways, including reviewing this work, appreciation, and encouragement. We also acknowledge ChatGPT for its assistance with proofreading and coding support.

\nocite{langley00}

\bibliography{example_paper}
\bibliographystyle{icml2026}

\newpage
\appendix
\onecolumn
\section{Appendix}

\subsection{Varying Strengths of RF and CF Across Distinct Mental States: An Illustrative Example}
This section illustrates how the strength of RF and CF inputs, and their interaction, varies across high-level perceptual processing and awake thought (imagination) states.\\
Read the ambiguous text in Figure 6 quickly at first, and then slowly \cite{kahneman2011thinking, selfridge1955pattern, rumelhart1986parallel}. When read quickly, our perception operates in a fast mode: automatic, pre-reflective, and anticipatory, neither too abstract nor too concrete \cite{parnas2021double, parnas2024phenomenological, blankenburg2001first}. We are able to focus on relevant information and decode it with a reasonable degree of confidence and coherence within that context. This corresponds to high-level perceptual processing \cite{Phillips2024cellular}: a state of being awake and conscious, characterized by basic, intuitive judgments sufficient for routine activities and everyday interactions, also known as common sense \cite{parnas2021double, parnas2024phenomenological, blankenburg2001first}. In this state, the strength and interaction of both RF and CF inputs range from moderate to high (Mod–High), mediated by cholinergic, noradrenergic, and orexinergic systems \cite{Phillips2024cellular}.\\
Now, if we slow down and take more time to interpret the information from different perspectives, we begin to perceive how identical characters can be interpreted differently depending on the context. For instance, consider the second character in both words in the first row, although visually identical, it is interpreted differently. Likewise, the first character in the second row and the second character in the third row are the same, yet each is perceived uniquely. This reflects a wakeful (imaginative) thought state \cite{Phillips2024cellular}: a heightened state of awareness and imagination that involves deep\footnote{Deep refers to the complex integration of diverse CFs at the cellular level, enabled by TPNs.}, slow, sequential, structured, deliberate, reflective, and well-justified judgments in context \cite{vlaev2018local}. In this case, the strength and interaction of RF and CF inputs range from high to maximal (High–Max), mediated by cholinergic, noradrenergic, and orexinergic pathways \cite{Phillips2024cellular}.
\\Overall, the combination of high-level perceptual processing and awake thought helps us navigate uncertainty and explore deeper, more nuanced, and less obvious meanings, moving beyond the literal interpretations offered by attention alone.
\begin{figure} 
	\centering
	\includegraphics[trim=0cm 0cm 0cm 0cm, clip=true, width=0.15\textwidth]{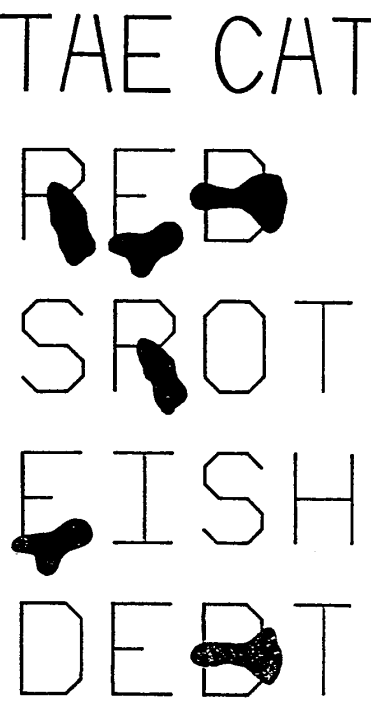}
	\caption{An example of ``thinking fast and slow" as discussed in \cite{kahneman2011thinking}, illustrates how solving a riddle can involve combining rapid, intuitive processing in high-level perceptual state (fast thinking) with more deliberate, reflective refinement in the awake thought state (slow thinking).} 
\vspace{-1.4em}
\end{figure}



\subsection{From $20^{\text{th}}$-century point neurons to $21^{\text{st}}$-century two-point neurons}

The brain is made up of billions of neurons, most receiving input from thousands of other neurons \cite{azevedo2009equal}. In $20^{th}$-century, neurally-inspired AI researchers assumed that neurons summed all their inputs to generate an output, a concept of point neurons (PNs), on which most current brain theories and AI systems are based \cite{hausser2001synaptic, lecun2015deep}. The neuroscience of the $21^{st}$ century \cite{larkum1999new, phillips2017cognitive, phillips2023cooperative, larkum2013cellular, major2013active, ramaswamy2015anatomy, larkum2022dendrites, adeel2020conscious, kording2000learning, SchumanAnnual, poirazi2020, larkum2018perspective, shine2016dynamics, shine2019human, shine2019neuromodulatory, shine2021computational, schulz2021gaba, kay2020contextual, kay2022comparison} has revealed that some neurons, particularly some pyramidal neurons, have two points of input integration, often referred to as TPNs. They combine information from the external environment (RF) at one point (basal dendrites) and internal world (CF) at another (apical dendrites). \\
Recent studies have concluded that the coupling of RF and CF is the hallmark of conscious processing \cite{aru2020cellular,  storm2024integrative, marvan2021apical}. This coupling does not occur under anesthesia (i.e., they are decoupled). In the conscious state, the gate within the cells is open that allows CF from a wide range of cortical and subcortical sources to propagate and evolve in the thalamocortical and corticocortical loops \cite{aru2020cellular}. The CF arriving from cortical and subcortical sources, and beyond, include feedback (FB) (reward), lateral intra-regional loops, and interactions between neocortical cells and the higher-order thalamus, and beyond \cite{Phillips2024cellular}. These contextual signals can be represented as FB, proximal (P), distal (D), and universal (U) \cite{adeel2020conscious, adeel2022unlocking, muckli2023cortical}. The coupling between CF and RF enables the propagation of coherent information streams. This does not occur in an unconscious state. \\
Coherent dendritic integration in TPNs occurs extensively across the cortical sheet, enhancing both individual neuronal complexity \cite{aru2020cellular, bachmann2020dendritic} and the capacity for moment-to-moment cooperation. This enables neurons to selectively transmit coherent thoughts while suppressing conflicting ones, ultimately promoting harmony and consistency in brain-wide activity \cite{phillips2023cooperative}.\\ Dysfunctional interactions between RF and CF inputs are linked to children with intellectual learning disabilities or significantly lower reasoning capabilities \cite{nelson2021dendritic, granato2024dysfunctions}. Patients with cognitive or perceptual conditions such as psychosis\footnote{The individuals whose ability to distinguish between imagination and reality is seriously impaired, for example,
believing that imagined voices are real \cite{phillips2003convergence, uhlhaas2006theory, phillips2013coherent}.} lack this balanced interaction between RF and CF, and struggle to focus on relevant information within a specific context \cite{pienkos2015intersubjectivity, parnas2021double, parnas2024phenomenological, sass2015faces, blankenburg2001first, zhu2020dark}. \\

\subsection{Extended Background}
Inspired by TPNs, computational neuroscientists have proposed various biologically plausible learning algorithms, including the recent burst-dependent synaptic plasticity (BDSP) \cite{payeur2021burst, Greedysingle}, which distinguishes between a single spike and a burst of spikes. It uses the top-down (apical) zone of the TPN to receive FB signals (bursting rate $b$) from higher perceptual levels as CF. The difference between the instantaneous and moving average of $b$ is multiplied by the event rate \textit{(e)} or RF to update weights \textit{(W)}. BDSP approximates loss-function gradients similar to backpropagation and performs comparably on complex image patterns, addressing the credit assignment problem effectively. However, BDSP and similar algorithms \cite{guerguiev2017towards, sacramento2018dendritic, IllingLocal, Greedysingle, zenke2017continual, kirkpatrick2017overcoming, kastellakis2016linking, bono2017modeling, limbacher2020emergence} focus mainly on apical inputs for learning. While their learning is TPN-inspired, their processing is not.\\
In contrast, there is ample neurobiological evidence suggesting that context serves as a modulatory factor \cite{marvan2024cellular, schulz2021gaba, kay2020contextual, kay2022comparison, phillips2023cooperative}. Neurophysiologists have proposed several biologically plausible asynchronous modulatory (MOD) transfer functions \cite{schulz2021gaba, kay2020contextual, kay2022comparison, phillips2023cooperative, schmid2023canonical} and their latest iterations \cite{pastorelli2023two, graham2025context}; however, whether there are many applications in which deep neural nets (DNNs) inspired by these transfer functions can outperform Point Neurons (PNs) \cite{adeel2022unlocking}-driven DNNs is yet to be seen. Below are these alternative well-established TPNs-inspired asynchronous MOD functions \cite{kay1998contextually, phillips2023cooperative}. In these modulatory transfer functions ($T_{M}$) eq (2-5), R is the driving force. 

\begin{equation}
T_{M1}(R, C) = \frac{1}{2}R(1+exp(RC))
\end{equation}
\begin{equation}
T_{M2}(R, C) = R+RC
\end{equation}
\begin{equation}
T_{M3}(R, C) = R(1+tanh(RC)) 
\end{equation}
\begin{equation}
T_{M4}(R, C) = R(2^{RC})
\end{equation}

Recently a new TPN MOD function \cite{adeel2022unlocking, muckli2023cortical, raza2024overlooked} incorporated not only FB from higher perceptual levels but also simultaneous events across hierarchies while processing FF information \cite{aru2020cellular, bachmann2020dendritic, shin2021memories, adeel2022unlocking, muckli2023cortical}. This MOD function uses CF to split the RF into coherent and incoherent streams at the cellular level, recombining only the coherent ones. The MOD function assigns greater weight to CF, amplifying or attenuating RF based on CF's strength. \\Recent computational results with convolutional neural nets (CNNs) \cite{adeel2022unlocking} have shown that TPNs-inspired CNN with this MOD function minimizes the transmission of large amounts of conflicting FF information to higher perceptual levels, greatly reducing (by orders of magnitude) the number of neurons required to process large amounts of heterogeneous real-world audio-visual data compared to PN-inspired CNNs \cite{adeel2022unlocking}. The most recent study in cellular psychology \cite{Phillips2024cellular} links these types of MOD functions to apical drive (AD), which is utilized in $\rm{Co}^4$.

\subsection{Bursting probabilities under distinct mental-state-dependent processing regimes}
\noindent\textbf{Slow-wave sleep/ AI and Apical Cooperation (AC):}

During SW sleep, apical input is functionally disconnected, and internal belief dynamics remain largely decoupled from sensory evidence. However, in this lowest-coherence regime, under low $R$ and low $C$ inputs, salience emerges only when both basal and apical inputs coincide, corresponding to minimal bursting probability in apical cooperation, such that:

\begin{equation}
\begin{aligned}
P^{LL}_2(b,a)
&= P_{1b}(b)\, P_{2a}(a) \\
&\equiv g^{LL}(b,a)
\end{aligned}
\end{equation}
Here, $P^{LL}_2(b,a)$ is the bursting probability under low basal ($b$) and low apical ($a$) inputs. $P_{1b}(b)$ is the probability of an initial somatic spike from basal input alone, and $P_{2a}(a)$ is the conditional contribution of apical input leading to a second spike (burst) \cite{graham2025context}. 

The reinterpreted MOD dynamics are defined as: $ \mathrm{MOD}(R, C) = g(R, C)$. 

\noindent\textbf{Apical Amplification:}

\begin{equation}
\begin{aligned}
P^{HL}_2(b,a)
&= P_{1b}(b)\, P_{2b}(b) \\
&\quad + P^{LL}_2(b,a)
\bigl[1 - P_{2b}(b)\bigr] \\
&\equiv f_b(b) + g^{HL}(b,a)
\end{aligned}
\end{equation}

$P^{HL}_2(b,a)$ represents bursting under high basal and low apical input. $P_{2b}(b)$ is the burst probability from basal input alone \cite{graham2025context}. 

The reinterpreted MOD dynamics are defined as: $ \mathrm{MOD}(R, C) = f(R) + g(R,C)$. 

\noindent\textbf{Apical Drive:}

\begin{equation}
\begin{aligned}
P^{LH}_2(b,a)
&= P^{H}_{2a}(a) \\
&\quad + P^{LL}_2(b,a)
\bigl[1 - P^{H}_{2a}(a)\bigr] \\
&\equiv f_a(a) + g^{LH}(b,a)
\end{aligned}
\end{equation}

$P^{LH}_2(b,a)$ is the bursting probability under low basal and high apical input. $P^{H}_{2a}(a)$ denotes the probability that apical input alone produces a spike \cite{graham2025context}. 

The reinterpreted MOD dynamics are defined as: $ \mathrm{MOD}(R, C) = f(C) + g(R,C)$. 

\noindent\textbf{Apical Drive + Awake:}

\begin{equation}
\begin{aligned}
P^{HH}_2(b,a)
&= P^{H}_{2a}(a) \\
&\quad + P^{HL}_2(b,a)
\bigl[1 - P^{H}_{2a}(a)\bigr] \\
&\equiv f_a(a) + f_b(b) + g^{HH}(b,a)
\end{aligned}
\end{equation}

$P^{HH}_2(b, a)$ is the bursting probability under high basal and high apical input. In this regime, both sources can independently initiate bursting, and their interaction further enhances amplification \cite{graham2025context}.

The reinterpreted MOD dynamics are defined as: $ \mathrm{MOD}(R, C) = f(R) + f(C) + g(R,C)$. 

All component functions $P^b_1(b)$, $P^a_2(a)$, $P^b_2(b)$, and $P^{H}_{2a}(a)$ are sigmoidal functions of stimulus amplitudes \cite{graham2025context}. 

\subsection{$\rm{Co}^4$ basic architecture}
$\rm{Co}^4$ architecture: $N$ denotes the number of input tokens, and each token has an embedding dimension of $E$. $Q_1$, $Q_2$,...,$Q_L$ represent the latent query tokens input to the associated Q-TPNs. $K_1$, $K_2$,...,$K_N$ represent the Key tokens input to the associated K-TPNs. $V_1$, $V_2$,...,$V_N$ represent the Value tokens input to the associated V-TPNs. This configuration forms part of the “seeing” state (i.e., sensory processing). In the “seeing as” state (i.e., perceptual and interpretive state), triadic modulation loops among questions ($Q$), clues (keys, $K$), and hypotheses (values, $V$) are executed through distal (D) and universal (U) contexts. Proximal (P) context represents normalization via information from neighboring neurons in the same population, including the prior information from the same neuron. The TPNs associated with $Q$, $K$, and $V$ are assumed to be analogous to three subtypes of pyramidal neurons, although their exact correspondence to neurobiologically distinguished subtypes is still under investigation. Through varying states of mind, high-level perceptual processing and wakeful thought, diverse, parallel reasoning chains are enabled. This mechanism incurs a computational cost of $\mathcal{O}(N \cdot L)$, where $L$ is a small fraction of the input length, making the overall cost approximately $\mathcal{O}(N)$. The triadic modulation loops, based on element-wise operations, add a nominal cost of $L \cdot N \cdot E$, which is significantly lower than that of the feedforward residual network used in standard Transformer blocks, a component $\rm{Co}^4$ does not require. $\rm{Co}^4$ can be viewed as a parallel, representation-level, silent yet deep form of Chain-of-Thought (CoT) reasoning \cite{wei2022chain} (a quiet mind), enabling multi-perspective inference without requiring sequential token-level generation, much like the brain’s cortico-thalamic modulation \cite{aru2020cellular, Phillips2024cellular, storm2024integrative}.
  \begin{figure}
    \centering
\includegraphics[width=\textwidth]{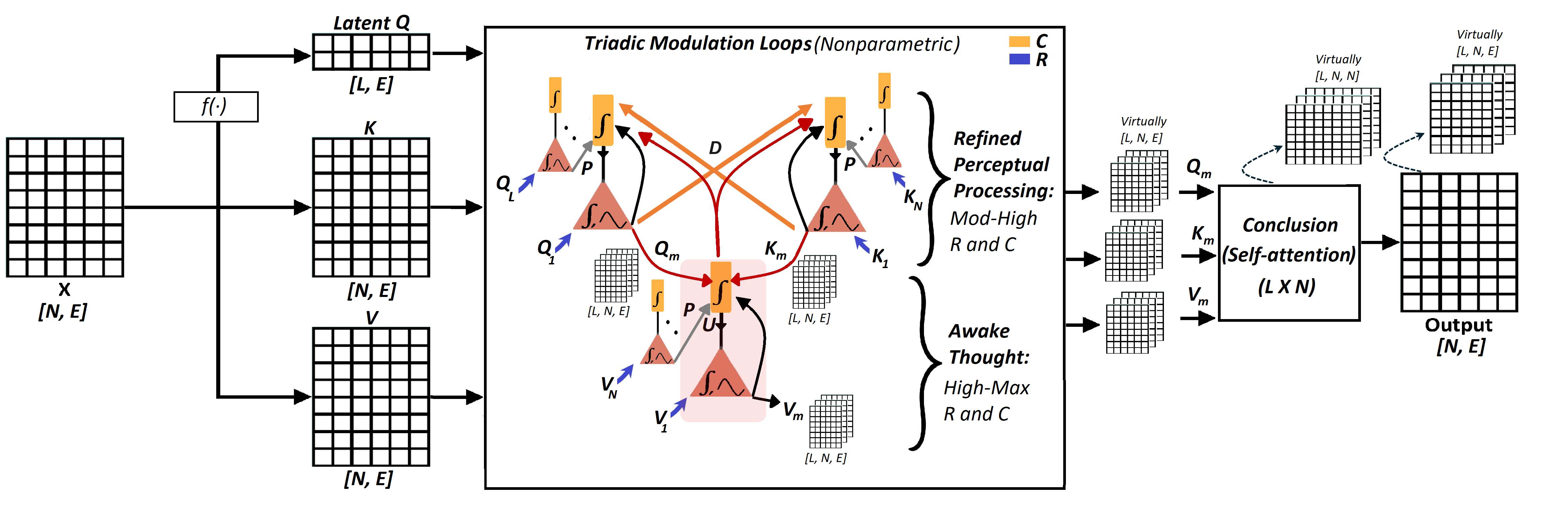}
    \caption{$\rm{Co}^4$ basic architecture: latent $Q$, $K$, and $V$ tokens evolve via TPN-like circuits and triadic modulation under distal, proximal, and universal context. Enables scalable, context-sensitive reasoning at an approximate cost of    
    $\mathcal{O}(N)$.}
    \label{fig:co4}
  \end{figure}
  \hfill

\subsection{Computational complexity}
Basic $\rm{Co}^4$ has a computational complexity of 
\[
\mathcal{O}(l \cdot N + \alpha)
\]
where \( N \) is the number of input tokens (patches or words), \( l \) is the number of latent tokens, and \( \alpha \) accounts for additional element-wise operations.  Instead of full attention between all \( N \) tokens,
\[
\Rightarrow \mathcal{O}(N^2),
\]
the model, similar to latent Transformers \cite{bi2024deepseek}, restricts this to \( l \times N \) interactions where \textit{l}
is a small fraction of the input length \textit{N} i.e., l $\ll N$,
\[
\Rightarrow \mathcal{O}(N \cdot l) \approx
\mathcal{O}(N)
\]
The element-wise operations in triadic modulation loops are significantly less expensive than matrix multiplications and lower than that of the FF residual network used in standard Transformer blocks, which $\rm{Co}^4$ does not require. $\rm{Co}^4$ scales linearly with input length, in contrast to the quadratic scaling in standard Transformers. For more explicit complexity analysis, the dominant MAC (multiply--accumulate) operations in standard Transformer model are approximated as:
\[
\textbf{MACs}_{\text{standard}} \approx L\Big(\,P\,E^2 + \,P^2\,E\Big)
\]
Where $E$, $P$, and $L$ represent the embedding dimension, number of tokens (or patches), and the number of layers, respectively. The term $\,P\,E^2$ represents the cost of the Q, K, V projections plus the FF network. The term $\,P^2\,E$ is due to the self--attention mechanism, which scales quadratically with $P$.\\
For the $\rm{Co}^4$ model, the dominant MAC operations are approximated as:
\[
\textbf{MACs}_{\text{$\rm{Co}^4$}} \approx L\Big(L_q\,E^2 + \,P\,E^2 + \,L_q\,P\,E\Big)
\]
$L_q$ represent the number of latent (query) tokens (with $L_q \ll P$). The term $L_q\,E^2$ arises from projecting the $L_q$ latent queries. The term $\,P\,E^2$ corresponds to the key and value projections. The term $\,L_q\,P\,E$ reflects the cost of  triadic modulation loops (emulating high-level perceptual processing and wakeful thought states), which scales linearly with $P$ since $L_q$ is small. 

For the scaled $\rm{Co}^4$, the dominant MAC operations can be approximated as
\begin{equation}
\text{MACs}_{\text{Co}^4} \approx L \left( N E^2 + N E + N \log k + k^2 E \right),
\end{equation}
where $k$ denotes the number of selected tokens in the Top-$k$ aggregation step. The term $N E^2$ corresponds to the projection cost for computing queries, keys, and values. The term $N E$ arises from the element-wise triadic modulation between the latent token and the patch tokens. The term $N \log k$ reflects the complexity of Top-$k$ token selection, which, under the constraint $k \sim N^{1/2}$, scales sub-quadratically and behaves linearly in practice due to the small constant $k$. Finally, the term $k^2 E$ accounts for the attention computation restricted to the reduced latent-enhanced token set, which remains negligible since $k \ll N$.

\vspace{-1em}

\subsection{Bursting probability dynamics across varying strengths of $R$ and $C$}

\begin{figure*}
	\centering
	\includegraphics[trim=0cm 0cm 0cm 0cm, clip=true, width=1\textwidth]{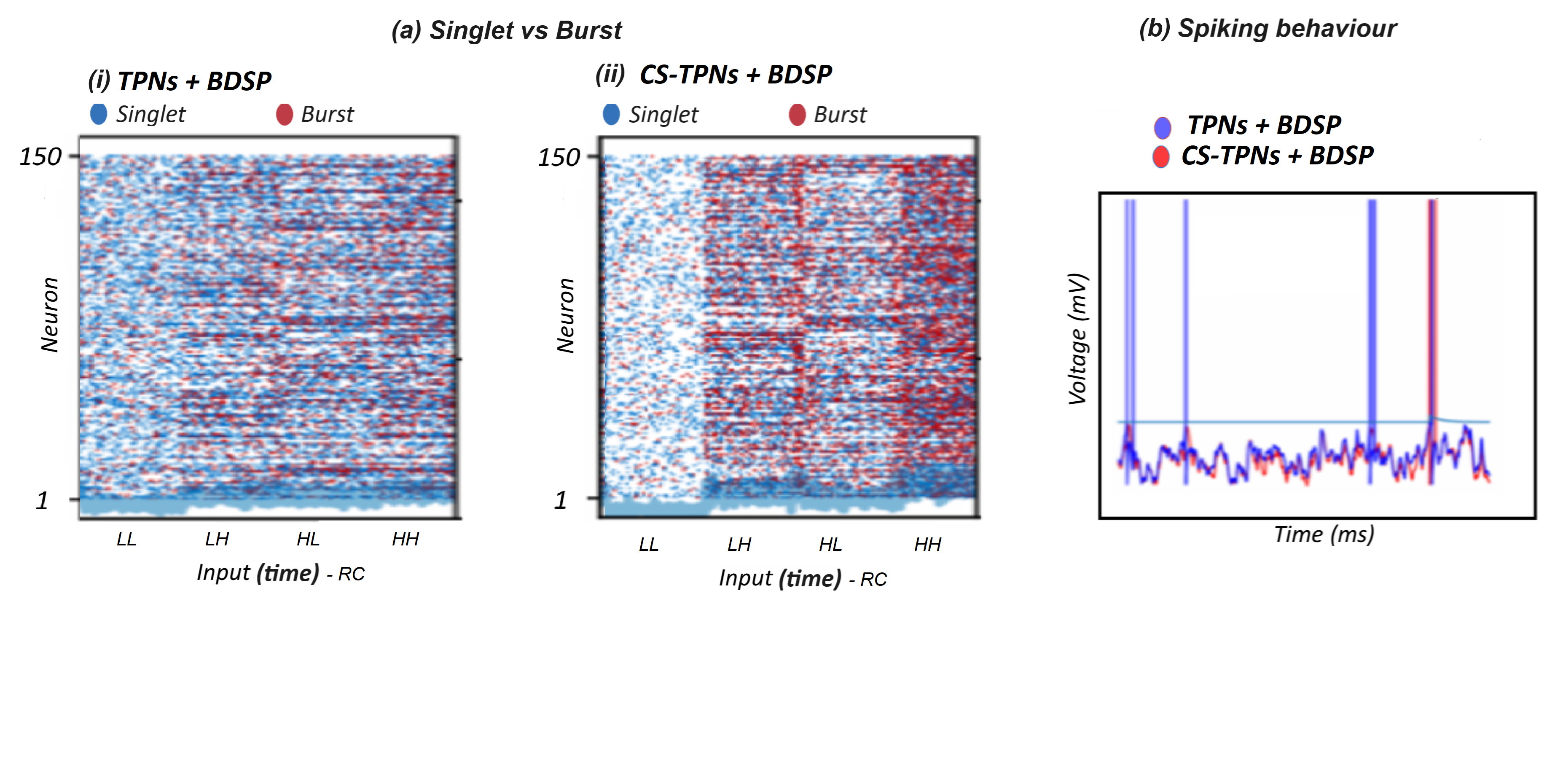}
	\caption{(a) Raster plots for 150 neurons. (i) standard TPNs + BDSP (ii) and context-sensitive (CS)-TPNs + BDSP with cooperative $MOD(R,C)$ dynamics. In CS-TPNS, a clear distinction in bursting probability is observed as the inputs transition from low–low (LL) to high–high (HH), corresponding to input combinations $00$, $01$, $10$, and $11$. (b) CS-TPNs fire substantially less frequently than TPNs alone and exhibit bursting primarily when $R$ and $C$ inputs are coherent.}
\vspace{-1.2em}
\end{figure*}

A variation of Eq. 12 was used in burst-dependent synaptic plasticity (BDSP) spiking simulations \cite{raza2024overlooked}. In these simulations, the strengths of the receptive ($R$) and contextual ($C$) inputs were varied across four regimes: low–low (LL), low–high (LH), high–low (HL), and high–high (HH). As shown in Fig. 8, the resulting bursting probability increases consistently as the system transitions from LL to HH, reflecting the increasing coherence between contextual and FF signals.

\begin{equation}
    Mod(I_s, I_c, I_u) = I_s + I_c (0.1 + |I_s|) + I_c I_u (2 + |I_s|)
    \label{Mod2}
\end{equation}

Here $I_s$, $I_c$, and $I_u$ denote the somatic (basal) current, the integrated contextual current, and the synergistic (universal) context current, respectively.

Following the formulation in \cite{payeur2021burst}, the somatic membrane potential dynamics and apical dendritic modulation of context-sensitive pyramidal neurons (CS-TPNs) can be described by the following simplified differential equations (Eqs. \ref{Vsoma}–\ref{wsoma}):

\begin{equation}
    \dot{V}_s = \frac{1}{\tau_s}(V_s - E_L) + \frac{1}{C_s}Mod(I_s,I_c,I_u) - \frac{1}{C_s}w_s
    \label{Vsoma}
\end{equation}

\begin{equation}
    \dot{w}_s = - \frac{1}{\tau_{ws}} w_s + b S(t)
    \label{wsoma}
\end{equation}

where $V_s$ represents the membrane potential of the somatic compartment, $\tau_s = 16\,\mathrm{ms}$ is the membrane time constant, $C_s = 370\,\mathrm{pF}$ is the membrane capacitance, and $E_L = -70\,\mathrm{mV}$ is the leak reversal potential. The variable $w_s$ denotes a spike-triggered adaptation current.

The total current applied to the soma is given by $Mod(I_s,I_c,I_u)$, which represents basal current $I_s$ modulated by contextual input $I_c$ together with synergistic interaction components $I_u$. The basal current $I_s$ corresponds to integrated synaptic inputs (including receptive excitatory and inhibitory inputs) together with basal noise. The contextual current $I_c$ reflects the combined influence of proximal and distal contextual signals as well as feedback error signals, together with dendritic noise. The synergistic term $I_u$ captures the extracted cooperative feedforward–contextual interaction components.

The adaptation variable $w_s$ evolves according to Eq. \ref{wsoma}, where $S(t)$ denotes the spike train of the neuron and $b = 200$ determines the strength of spike-triggered adaptation. A spike is emitted whenever $V_s$ crosses a dynamic threshold of $-50\,\mathrm{mV}$. Following a spike, the threshold is increased by $2\,\mathrm{mV}$ and decays back to $-50\,\mathrm{mV}$ with a time constant of $27\,\mathrm{ms}$. After each spike, the membrane potential is reset to the resting voltage $V_r = -70\,\mathrm{mV}$. These parameter values follow those reported in \cite{payeur2021burst}.

A more detailed analysis of these spiking simulations can be found in \cite{raza2024overlooked}.

\subsection{Significance Statement and Future Directions}
While the development of human-like AI raises important ethical considerations, the evidence presented here suggests that awake imaginative thought may be grounded in cooperative neural processes shaped by evolution \cite{bregman2020humankind, phillips2023cooperative}. In addition to demonstrating fast and economical learning by emulating awake imaginative regimes, the present investigation suggests that context is actively constructed from available information to support internal hypotheses. When early internal inferences are misleading, convergence to accurate conclusions may require substantially more time or may fail altogether under constrained computational resources. This motivates the hypothesis that effective awareness, in a computational sense, arises from the capacity to iteratively evolve coherent latent queries, clues, and hypotheses in alignment with incoming evidence. Without such alignment, perceptual representations may become unstable or misleading.

These observations raise a broader question: whether the cellular mechanisms underlying context-sensitive integration in higher mammals can inform principled models of common-sense and imaginative reasoning, and thereby guide the design of future AI systems that are more ethical, efficient, robust, and context-aware. Overall, such biologically grounded mechanisms may help advance AI beyond conventional passive information processing toward more coherent, context-sensitive, and real understanding grounded in internal–external representational alignment.



\end{document}